\documentclass[sigconf]{acmart_}

\usepackage{multirow}
\usepackage{makecell}
\usepackage{tabularx}
\usepackage{amsmath}
\usepackage{algorithm}
\usepackage{algpseudocode}

\usepackage{subcaption}
\usepackage{graphicx}
\usepackage[table]{xcolor}
\usepackage[dvipsnames]{xcolor}

\AtBeginDocument{%
  }

\copyrightyear{2026}
\acmYear{2026}
\acmDOI{XXXXXXX.XXXXXXX}
\acmConference[Conference acronym 'XX]{Make sure to enter the correct
  conference title from your rights confirmation email}{June 03--05,
  2018}{Woodstock, NY}
\acmBooktitle{Woodstock '18: ACM Symposium on Neural Gaze Detection,
 June 03--05, 2018, Woodstock, NY}
\acmISBN{978-1-4503-XXXX-X/2018/06}

\begin{document}

\title{UniCast: A Unified Framework for Instance-Conditioned Multimodal Time-Series Forecasting}

%
\author{Sehyuk Park}
\email{sehyuk.park@student.unimelb.edu.au}
\affiliation{%
  \institution{The Universito of Melbourne}
  \city{Melbourne}
  \state{Victoria}
  \country{Australia}
}
\author{Soyeon Caren Han}
\email{caren.han@unimelb.edu.au}
\affiliation{%
  \institution{The Universito of Melbourne}
  \city{Melbourne}
  \state{Victoria}
  \country{Australia}
}
\author{Eduard Hovy}
\email{eduard.hovy@unimelb.edu.au}
\affiliation{%
  \institution{The Universito of Melbourne}
  \city{Melbourne}
  \state{Victoria}
  \country{Australia}
}

%
\renewcommand{\shortauthors}{Park et al.}

\begin{abstract}
Time series forecasting underpins applications in finance, healthcare, and environmental monitoring. Despite the success of Time Series Foundation Models (TSFMs), existing approaches operate in a unimodal setting and rely on static prompts or fixed fusion schemes, limiting their ability to exploit multimodal context and adapt to instance-level variation.
We propose UniCast, a parameter-efficient multimodal framework that extends TSFMs through instance conditioned prompting and dynamic modality routing. UniCast infers a conditional prompt from time series, vision, and text inputs via a Transformer-based contextual distiller, enabling input-specific adaptation without updating the forecasting backbone. To regulate how auxiliary modalities influence predictions, UniCast employs Modality Routing, a cross-attention mechanism that estimates modality relevance given the current temporal state and selectively amplifies informative signals while suppressing noise.
Integrated with a frozen TSFM via soft prompt tuning, UniCast preserves foundation-level generalization while enabling effective multimodal control. Extensive experiments across diverse forecasting benchmarks show that UniCast consistently outperforms all existing TSFM baselines, demonstrating that instance-conditioned multimodal control is critical for next-generation time series forecasting.
\end{abstract}




\maketitle


\section{Introduction}

\begin{figure}
    \centering
    \includegraphics[width=\columnwidth]{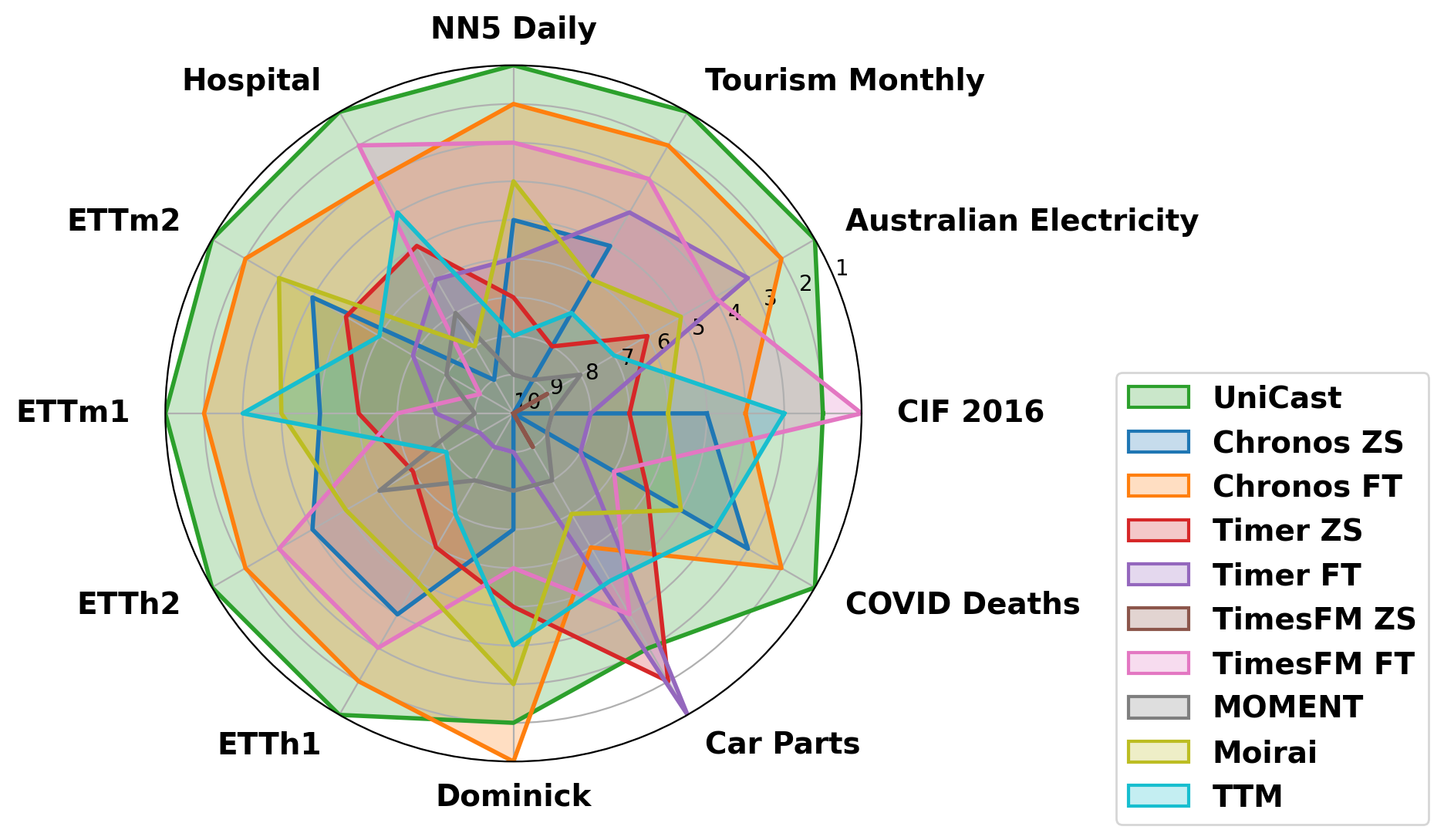}
    \caption{Radar plot showing per-dataset model rankings (1 = best) comparing UniCast (Chronos) with strong time-series foundation model baselines under zero-shot (ZS) and full fine-tuned (FT) settings. UniCast consistently attains top/near-top ranks across heterogeneous domains, highlighting the benefit of instance-conditioned multimodal control over static or unimodal forecasting approaches.}
    \label{fig:radar}
    \vspace{-0.3cm}
\end{figure}

Recent time series forecasting research has been driven by \emph{Time Series Foundation Models} (TSFMs), which leverage large-scale pretraining to learn transferable temporal representations and achieve strong zero-shot and few-shot generalization across diverse forecasting tasks~\cite{ansari2024chronos, liu2024timer}.
By capturing generic temporal dynamics, TSFMs reduce the need for task-specific modeling and extensive feature engineering.
Despite their success, existing TSFMs operate almost exclusively in a \emph{unimodal} regime, treating time series as isolated numerical signals. In real-world settings, however, forecasting problems are rarely unimodal. Temporal observations are often accompanied by rich auxiliary context, such as visual information (e.g., plots, sensor imagery, or environmental snapshots) and textual descriptions (e.g., metadata, annotations, or event summaries). These modalities frequently provide complementary signals that disambiguate temporal patterns, improve robustness, and support generalization under
distribution shift. Ignoring such context fundamentally limits the expressive power of current TSFMs.

A natural response is to incorporate multimodal information into forecasting models. Prior work has explored modality augmentation through language-based prompting and multimodal foundation models~\cite{jin2023timellm, wang2025chattime}. However, most existing approaches rely on \emph{static prompts} or \emph{fixed fusion schemes}, implicitly assuming that auxiliary modalities are uniformly informative across all instances. This assumption is problematic in practice: the relevance of vision or text varies substantially depending on the temporal state, noise level, and data regime, and indiscriminate fusion can introduce spurious correlations or amplify irrelevant signals.

We argue that the central challenge in multimodal time series forecasting is not merely \emph{how} to fuse modalities, but \emph{when} and \emph{to what extent} each modality should influence prediction. From this perspective, multimodal forecasting is fundamentally a problem of \emph{instance-level modality relevance identification}. Existing TSFMs lack architectural mechanisms to represent external contextual signals and to regulate their influence in an input-dependent manner. As a result, identical temporal patterns which were observed under different contextual conditions are treated as equivalent, leading to brittle behavior and degraded performance in challenging scenarios.

To address this limitation, we propose \textbf{UniCast}, a unified and
parameter-efficient multimodal prompting framework that extends pretrained TSFMs through \emph{instance-conditioned adaptation} and \emph{dynamic modality control}.
UniCast is designed around two complementary components.
First, \emph{Conditional Prompting} infers instance-conditioned contextual representations from time series, vision, and text inputs using a lightweight Transformer-based context distiller. 
These contextual representations capture modality-aware information relevant to the current input and adapt the forecasting process without modifying the TSFM backbone.
Second, \emph{Modality Routing} dynamically regulates how auxiliary modalities influence prediction. Using cross-attention conditioned on the evolving temporal representation, UniCast selectively amplifies informative modalities while suppressing noise, enabling fine-grained and interpretable control over multimodal contributions.

Crucially, UniCast preserves the generalization strengths of foundation models by keeping all pretrained encoders and the TSFM backbone frozen. Trainable parameters are confined to prompt generators, routing layers, and lightweight projection modules, enabling parameter-efficient multimodal adaptation without retraining large models or sacrificing scalability. We evaluate UniCast on a diverse set of time series forecasting benchmarks spanning multiple domains and data regimes.
Extensive experiments show that UniCast consistently outperforms strong TSFM baselines and fine-tuned variants, particularly in low-signal and distribution-shift settings. These results demonstrate that \emph{instance-conditioned multimodal control}, rather than static fusion, is critical for next-generation time series forecasting.

In summary, this paper makes the following contributions:
\begin{itemize}
    \item We identify instance-level modality relevance as a key missing capability in existing TSFMs and formalize multimodal forecasting as a problem of adaptive contextual control.
    \item We introduce UniCast, a parameter-efficient multimodal framework that extends frozen TSFMs through conditional prompting and modality routing.
    \item We provide comprehensive empirical evidence showing that dynamic, instance-conditioned multimodal integration substantially improves forecasting performance.
\end{itemize}

\section{Related Work}

\noindent\textbf{Foundation Models for Time Series Forecasting.}
Recent Time Series Foundation Models (TSFMs), such as Timer\cite{liu2024timer}, TimesFM\cite{das2024timesfm}, TTM\cite{ekambaram2024ttm}, Chronos\cite{ansari2024chronos}, MOMENT\cite{goswami2024moment}, and Moirai\cite{woo2024moirai}, are pretrained on large-scale time-series corpora and have demonstrated strong generalization across domains and downstream forecasting tasks.
By learning generic temporal dynamics, these models capture robust inductive biases for numerical time-series patterns.
However, TSFMs are inherently unimodal: they operate exclusively on historical numerical observations.
As a result, identical temporal patterns occurring across different external contexts, such as domain conditions, environments, or events, are treated as equivalent, despite potentially having different semantic meanings. This limitation is not merely due to missing information; it arises from a structural constraint: the input and representation spaces of TSFMs lack mechanisms to represent, condition on, or reason about external contextual signals.
Recent studies have shown that incorporating contextual information from auxiliary modalities can substantially improve forecasting performance, yet existing TSFMs are designed to forgo this capability.

\noindent\textbf{Modality-Augmented Time Series Forecasting.}
To enrich forecasting with contextual information, prior work has explored incorporating auxiliary modalities such as language and vision. \emph{LLM-based approaches} leverage pretrained language models to inject external knowledge and reasoning capabilities into time-series forecasting.
Methods such as LSTPrompt\cite{liu2024lstprompt}, Time-LLM\cite{jin2023timellm}, S2IP-LLM\cite{pan2024s2ip}, TEST\cite{sun2023test}, LangTime\cite{niu2025langtime}, TimeXL\cite{jiang2025timexl}, ChatTime\cite{wang2025chattime}, and CALF\cite{liu2025calf} convert time-series inputs into textual representations or combine them with text prompts.
While effective in certain settings, these approaches suffer from modality mismatch: transforming continuous time-series into discrete tokens often flattens local temporal dynamics and reduces temporal resolution. Moreover, early fusion of textual and time-series representations entangles modality contributions, making it difficult to regulate or analyze the influence of each modality at the instance level during prediction.
\emph{Vision-based approaches} leverage pretrained vision encoders to transform time-series data into images or visual patterns, as in VisionTS\cite{chen2025visionts}, ViTime\cite{yang2025vitime}, and DMMV\cite{shen2025dmmv}.
Although these methods benefit from powerful visual pattern recognition, the transformation from time to image can obscure temporal ordering and causal structure.
Visually similar patterns do not necessarily correspond to decision-relevant temporal dependencies, limiting the fidelity with which visual features reflect forecasting-critical dynamics.
Several recent works attempt explicit cross-modal alignment.
TimeCMA\cite{liu2024timecma} performs alignment between time-series and text representations, but requires training the time-series encoder and decoder from scratch, preventing the use of generalization priors learned by pretrained TSFMs.
Time-VLM\cite{zhong2025timevlm} leverages vision–language models to provide multimodal context for forecasting; however, by projecting vision and language into a joint embedding space, modality contributions become tightly coupled, making instance-level control or selective reliance difficult.
More broadly, existing multimodal approaches do not support routing mechanisms that dynamically emphasize or suppress modalities based on input-specific relevance.

\noindent\textbf{Positioning and Contributions.}
Our work addresses three key gaps in existing literature.
First, a \emph{context integration gap}: pretrained TSFMs lack architectural support for representing or conditioning on external contextual signals.
Second, a \emph{controllability gap}: prior multimodal approaches rely on joint or early fusion, limiting the ability to disentangle, regulate, or analyze modality contributions at the instance level.
Third, an \emph{efficiency gap}: many methods require training time-series models from scratch or fine-tuning entire multimodal backbones, reducing parameter efficiency and scalability.

UniCast bridges these gaps by retaining a frozen pretrained TSFM as the forecasting backbone while introducing input conditioned multimodal control. Through conditional prompting and modality routing, UniCast learns \emph{when and how much to rely on each modality}, rather than fusing them indiscriminately.
This design enables parameter-efficient multimodal adaptation while preserving the temporal inductive biases and generalization strengths of foundation time-series models.

\begin{figure*}[t]
\centering
\includegraphics[width=0.85\textwidth]{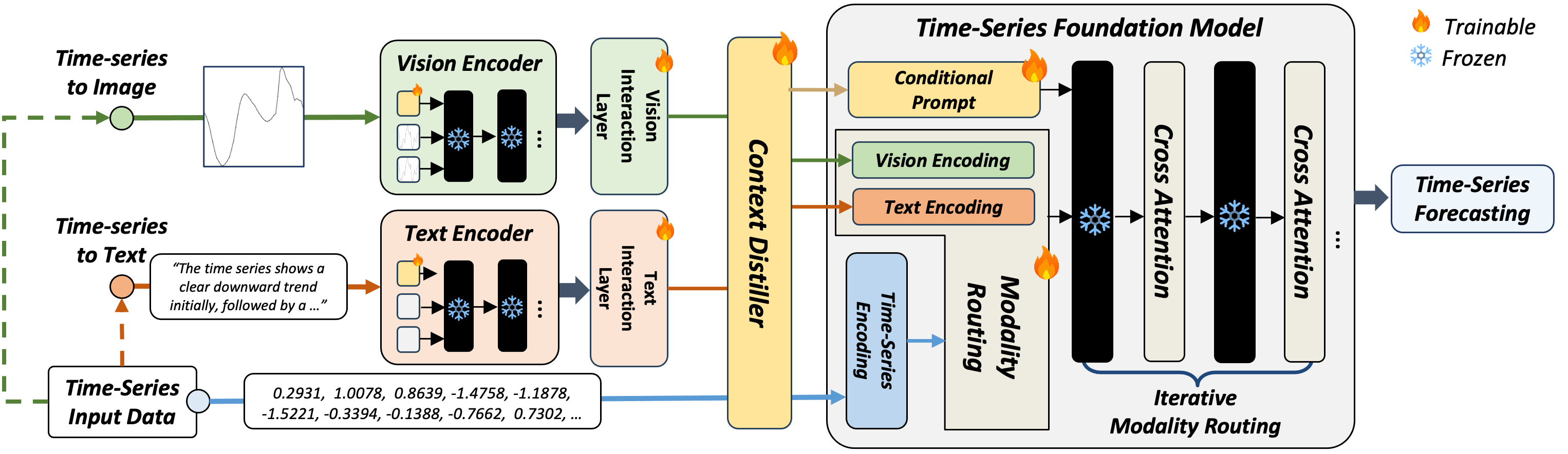}
\caption[UniCast Overview]{UniCast formulates multimodal time-series forecasting as an instance-level modality credit assignment problem. (Left) Time series, vision, and text inputs are encoded by frozen pretrained backbones to preserve foundation-level generalization. (Middle) Conditional Prompting distills cross-modal context to generate instance-specific soft prompts, enabling adaptive conditioning under temporal non-stationarity. (Right) The Modality Routing mechanism performs iterative, input-dependent credit assignment via cross-attention, selectively injecting informative auxiliary signals into the TSFM while suppressing irrelevant or noisy modalities during forecasting.}
\label{fig:model_architecture}
\end{figure*}

\section{Method}
\subsection{Design Philosophy}
We frame multimodal time-series forecasting as a problem of instance-level modality relevance identification. While incorporating auxiliary modalities can improve forecasting, treating all modalities as equally informative often leads to brittle behavior in real-world settings, where modality relevance varies across instances and over time.
Most existing approaches\cite{cao2023tempo, jin2023timellm, liu2025dp} rely on static prompts or heuristic fusion schemes. Static prompts encode a fixed contextual prior derived from the training distribution, limiting adaptation to temporal non-stationarity at inference time. Likewise, heuristic fusion methods impose fixed integration patterns that implicitly assume uniform modality importance, even though a modality may be informative in some instances and distracting in others. These limitations stem from the inability to assess modality-specific information gain conditioned on the current input.

Motivated by adaptive sensory modulation in biological systems, we propose a framework that explicitly separates context inference from modality utilization. This design yields a two-stage architecture consisting of Conditional Prompting, which infers instance-specific contextual priors, and Modality Routing, which adaptively regulates how multimodal information is integrated into the forecasting backbone.

\subsection{UniCast}
Following this philosophy, \emph{UniCast} operates on multimodal representations extracted from frozen pretrained vision and text encoders, together with a frozen time-series foundation model (TSFM). Rather than directly injecting static or modality-specific prompts, UniCast decomposes multimodal adaptation into two explicitly separated stages. From a parameter-efficiency perspective, UniCast can be viewed as a \emph{multimodal soft-prompt-based PEFT} framework, where all pretrained encoders and the TSFM backbone remain frozen, and adaptation is achieved solely through lightweight prompt generators and routing layers.

First, \emph{Conditional Prompting} observes time-series, vision, and text inputs to infer an \emph{instance-conditioned contextual prompt} that captures the relevant context of the current input. This prompt serves as a contextual prior that adapts the interpretation of multimodal signals without modifying the forecasting backbone.
Second, \emph{Modality Routing} performs \emph{input-dependent credit assignment}, dynamically estimating the relevance of auxiliary modalities conditioned on the current temporal state and selectively integrating informative signals into the TSFM.
All pretrained encoders remain frozen during training, with learnable parameters restricted to the conditional prompt generators, modality routing layers, and lightweight projection modules. This design enables parameter-efficient multimodal adaptation while preserving the generalization capabilities of the foundation model.

\subsection{Problem Setup and Inputs}
We consider a multimodal time-series forecasting setting where each input instance consists of a target time series segment
$X \in \mathbb{R}^{T \times D}$, together with optional auxiliary modalities including a visual input $V$ and a textual description $L$.
Given historical observations of the target series, the forecasting objective is to predict future values over a horizon $H$.
The time-series input $X$ is transformed into patch-level embeddings following standard preprocessing used in TSFMs.
Visual and textual inputs are encoded using frozen pretrained vision and text encoders to obtain contextual embeddings.
UniCast operates exclusively on these intermediate representations, enabling multimodal adaptation without modifying the underlying encoders or the forecasting backbone. Unless otherwise specified, all auxiliary modalities are assumed to be aligned with the input time-series segment. The complete computation flow of UniCast is summarized in Algorithm~\ref{alg:algorithm} at Appendix~\ref{app:algorithm}.

\subsection{Conditional Prompting}
Static prompts encode a fixed contextual prior and therefore cannot adapt to temporal non-stationarity or instance-level variation at inference time.
To overcome this limitation, we formulate the prompt as an \emph{input-conditioned contextual representation} inferred from multimodal inputs. 
Importantly, conditional prompting does not directly inject multimodal signals into the forecasting backbone; instead, it modulates the interpretation of multimodal information that is selectively utilized downstream.
Inferring such a representation from heterogeneous modalities requires a mechanism that can model cross-modal interactions without imposing fixed assumptions on modality importance.

Conditional prompt inference is performed in two stages.
\textbf{(1) Modality-aware context distillation.}
We extract token-level embeddings from frozen vision and text encoders, and patch-level embeddings from the time-series input. For each modality, a transformer-based context distiller is applied to extract an instance-specific contextual representation.
Rather than enforcing a shared aligned space across modalities, the context distillation process operates in a modality-aware manner, producing contextual signals that are subsequently used for prompt generation and modality routing.
The resulting contextual representations are used only for conditional prompt inference and are not directly consumed by the forecasting backbone.
\textbf{(2) Prompt generation.}
The distilled contextual representations are combined with pooled time-series patch embeddings to generate the final conditional prompt.
Specifically, the first token of the last hidden state is used as the vision contextual embedding, while average pooling is applied to the text embeddings.
The concatenated representation is processed by a Transformer encoder to produce an instance-specific conditional prompt that captures both temporal dynamics and multimodal context.
Alternative strategies for constructing contextual embeddings are analyzed and shared in Appendix~\ref{app:representationstrategy}.

\subsection{Modality Routing}
Heuristic fusion methods assign fixed importance to each modality and therefore cannot adapt to instance-level variation in modality relevance.
In contrast, \emph{Modality Routing} explicitly performs \emph{input-conditioned credit assignment}, regulating how much auxiliary information is injected into the forecasting backbone.
We implement Modality Routing using a cross-attention mechanism in which time-series patch embeddings act as queries, while contextual embeddings from vision and text modalities serve as keys and values.
This design evaluates modality relevance in the context of current temporal dynamics and selectively amplifies informative signals while suppressing noise.
Modality relevance is computed as
\begin{equation}
\alpha_{t}^{(m)} = \mathrm{softmax}\left(
    \frac{Q_t K_m^\top}{\sqrt{d}}
\right),
\label{eq:modality_routing}
\end{equation}
where $Q_t$ denotes the time-series representation at temporal position $t$, 
$K_m$ denotes the contextual embedding of modality $m$, 
and $\alpha_{t}^{(m)}$ represents the instance-level relevance (credit) assigned to modality $m$.
The resulting cross-attention output is injected into the time-series representations and iteratively refined through the TSFM, allowing modality relevance to be re-evaluated as temporal representations evolve.
All routing layers share weights for parameter efficiency and consistency.
Unlike standard cross-attention used for joint representation learning, modality routing is explicitly designed for relevance estimation and selective injection, with attention weights interpreted as modality credit assignment rather than alignment scores.

\section{Evaluation Setup}
\label{subsec:datasets}

\begin{table}[t]
  \centering
  \caption[Evaluation Datasets]{Summary of evaluation datasets used in our experiments, including data frequency, context length ($C$), forecast length ($H$), and \# of Series. Detailed descriptions of each dataset and their sources are provided in the Appendix~\ref{app:datasetdetail}.}
  \label{tab:evaluationdatasets}
{\fontsize{9}{11}\selectfont
  \begin{tabularx}{\linewidth}{Xcccc}
    \toprule
    \textbf{Dataset} & \textbf{Frequency} & \textbf{\makecell{$C$}} & \textbf{\makecell{$H$}} & \textbf{\# of Series}\\
    \midrule
    NN5 Daily & 1 Day & 56 & 56 & 111\\
    Australian Electricity & 30 Minutes & 48 & 48 & 5\\
    Tourism Monthly & 1 Month & 24 & 24 & 366\\
    COVID-19 Deaths & 1 Day & 30 & 30 & 266\\
    Car Parts & 1 Month & 12 & 12 & 2674\\
    CIF 2016 & 1 Month & 12 & 12 & 72\\
    Dominick & 1 Week & 8 & 8 & 100014\\
    Hospital & 1 Month & 12 & 12 & 767\\
    ETTh & 1 Hour & 24 & 24 & 14\\
    ETTm & 15 Minutes & 96 & 24 & 14 \\
    \bottomrule
  \end{tabularx}
}
  \vskip -0.1in
\end{table}

\subsection{Datasets and Metrics\protect\footnotemark}
\footnotetext{More details, including dataset statistics, data partitioning, normalization and preprocessing, are provided in Appendix~\ref{app:datasetdetail}.}
Following the zero-shot evaluation protocol of Chronos~\cite{ansari2024chronos}, we evaluate UniCast on a representative subset of Benchmark II datasets that span diverse domains, frequencies, and data regimes. This selection is designed to cover varying levels of temporal signal strength and contextual availability,  allowing us to assess the effectiveness of multimodal, parameter-efficient adaptation under heterogeneous conditions.
The selected datasets exhibit varying degrees of auxiliary information availability across textual and visual modalities,  ranging from rich semantic descriptions to minimal metadata, and from structured visual patterns to stochastic noise. This diversity enables us to evaluate whether UniCast can selectively leverage informative multimodal cues while suppressing irrelevant signals.
To ensure a fair comparison, we exclusively evaluate on test sets that were not included in the pre-training phase of Chronos. This setting prevents information leakage and allows us to assess generalization to unseen domains,  rather than performance driven by memorization.
Beyond achieving a new state-of-the-art (SOTA), this experimental design serves to provide a theoretical and empirical basis for under which data conditions multimodal synergy and PEFT are most effective. Consequently, our findings offer practical guidelines for deploying multimodal time-series models in real-world industrial applications.
Following standard time-series forecasting practice, we use Mean Squared Error (MSE) as our primary evaluation metric to quantify prediction accuracy.
Importantly, all evaluations are conducted under a parameter-efficient setting, where the pretrained backbones remain frozen and only lightweight prompt and routing modules are trained. This allows us to isolate the performance gains attributable to multimodal integration and PEFT-based adaptation.

\begin{table*}[t]
    \caption[Full Forecasting Performance]{Full forecasting performance comparison across twelve datasets. We report the Mean Squared Error (MSE) of UniCast and six strong time-series foundation model baselines, along with fine-tuned results for three models (Chronos, Timer, and TimesFM). UniCast significantly outperforms all baselines on average, with UniCast (Chronos, CLIP, Qwen) and UniCast (Chronos, BLIP, LLaMA) achieving the lowest and second-lowest MSE, respectively. \textbf{Bold} indicates the best performance and \underline{underlined} the second best for each dataset. $^*$ indicates Timer was pre-trained on Australian Electricity dataset.}
    \label{tab:mainresult}
{\fontsize{9}{11}\selectfont
    \setlength{\tabcolsep}{4pt}
    \begin{tabularx}{\linewidth}{>{\centering\arraybackslash}X| cccccccccccc |c}
    \hline
    \multirow{2}{*}{} 
        & \multirow{2}{*}{\textbf{NN5}}
        & \multirow{2}{*}{\textbf{Aus}}
        & \multirow{2}{*}{\textbf{Tour}}
        & \multirow{2}{*}{\textbf{COV}}
        & \multirow{2}{*}{\textbf{Car}}
        & \multirow{2}{*}{\textbf{CIF}}
        & \multirow{2}{*}{\textbf{Dom}}
        & \multirow{2}{*}{\textbf{Hos}}
        & \multicolumn{4}{c|}{\textbf{ETT}}
        & \multirow{2}{*}{\textbf{Avg}} \\
    \cline{10-13}
        & & & & & & & & 
        & \textbf{h1} & \textbf{h2} & \textbf{m1} & \textbf{m2} & \\
    \hline
    \hline
    \textbf{Chronos}$^{\spadesuit}$ & 0.7913 & 2.6701 & 2.4682 & 0.1562 & 1.3028 & 7.4613 & 1.3396 & 2.6128 & 0.2891 & 0.1718 & 0.1278 & 0.0828 & 1.6228\\
    \textbf{Chronos}$^{\heartsuit}$ & \underline{0.4635} & 0.3606 & 1.3896 & 0.1197 & 0.9957 & 7.3853 & \textbf{0.9453} & 2.3626 & \underline{0.2661} & 0.1390 & 0.1022 & 0.0357 & 1.2138\\
    \textbf{Timer}$^{\spadesuit}$ & 1.0986 & 1.4620$^*$ & 2.6216 & 0.2178 & 0.9573 & 7.5969 & 1.1943 & 2.4551 & 0.3149 & 0.2821 & 0.1390 & 0.0837 & 1.5353\\
    \textbf{Timer}$^{\heartsuit}$ & 0.9368 & 0.4839$^*$ & 2.4492 & 0.3846 & \textbf{0.9197} & 7.7260 & 2.1475 & 2.4697 & 0.6712 & 0.6476 & 0.2410 & 0.2203 & 1.6081\\
    \textbf{TimesFM}$^{\spadesuit}$ & 1.7585 & 2.2816 & 3.1699 & 1.2082 & 1.0730 & 8.0750 & 3.7334 & 2.6340 & 1.8952 & 1.8955 & 1.9977 & 2.0153 & 2.6448\\
    \textbf{TimesFM}$^{\heartsuit}$ & 0.5199 & 0.5163 & 1.5287 & 0.3584 & 0.9846 & \textbf{7.2993} & 1.3271 & 2.3492 & 0.2817 & 0.1410 & 0.1893 & 0.4376 & 1.3278\\
    \textbf{MOMENT}$^{\spadesuit}$ & 1.1302 & 1.5763 & 2.6899 & 0.5173 & 1.0524 & 7.7479 & 2.0257 & 2.5624 & 0.3684 & 0.2748 & 0.2446 & 0.2222 & 1.7010\\
    \textbf{Moirai}$^{\spadesuit}$ & 0.5294 & 1.1993 & 2.5958 & 0.1890 & 1.0459 & 7.4979 & 1.0567 & 2.5741 & 0.3070 & 0.2428 & 0.1224 & 0.0628 & 1.4519\\
    \textbf{TTM}$^{\spadesuit}$ & 1.1172 & 1.5117 & 2.6022 & 0.1756 & 0.9913 & 7.3839 & 1.0741 & 2.4479 & 0.3546 & 0.3576 & 0.1183 & 0.1213 & 1.5213\\
    \hline
    \textbf{UniCast(Ours)} &&&&&&&&&&&&&\\
    w/ CLIP,Qwen & \underline{0.4635} & \underline{0.3431} & \underline{1.3573} & \textbf{0.1066} & \underline{0.9525} & 7.3802 & \underline{0.9851} & \underline{2.3216} & 0.2690 & \underline{0.1358} & \textbf{0.0951} & 0.0307 & \textbf{1.2034}\\
    w/ BLIP,LLaMA & \textbf{0.4577} & \textbf{0.3059} & \textbf{1.3467} & \underline{0.1095} & 0.9766 & \underline{7.3765} & 1.0476 & \textbf{2.3187} & \textbf{0.2639} & \textbf{0.1297} & \underline{0.0997} & \textbf{0.0293} & \underline{1.2052}\\
    \hline
    \end{tabularx}
}
\begin{flushleft}
\footnotesize
$^{\spadesuit}$ Zero-shot. $^{\heartsuit}$ Fine-tuned.
\end{flushleft}
\end{table*}

\subsection{Baselines}
To comprehensively evaluate the effectiveness of our proposed framework, we compare UniCast against a wide range of state-of-the-art time series models. These baselines were selected based on their architectural diversity and proven zero-shot forecasting performance.
We consider the following six models as our primary baselines. Notably, to directly quantify the performance gain provided by our multimodal integration, we adopt Chronos, Timer, and TimesFM not only as standalone baselines but also as the underlying backbones for our framework.

\begin{itemize}
    \item Chronos\cite{ansari2024chronos}: A T5-based architecture pretrained on quantized and scaled time series data treated as discrete tokens. Chronos leverages time series-specific data augmentations to improve generalization in zero-shot and few-shot forecasting scenarios.
    \item Timer\cite{liu2024timer}: A decoder-only autoregressive model trained via next-token prediction on large-scale time series data. It demonstrates strong adaptability to diverse forecasting tasks through causal modeling.
    \item TimesFM\cite{das2024timesfm}: A decoder-only foundation model for time-series forecasting, pretrained on large-scale temporal data using causal attention, enabling strong zero-shot performance across diverse domains.
    \item MOMENT\cite{goswami2024moment}: A family of transformer-based models pre-trained with a masked time-series prediction objective that enables general time-series analysis.
    \item MOIRAI\cite{woo2024moirai}: A universal forecasting model built with a Masked Encoder-based Transformer to handle diverse frequencies and variables in a single framework.
    \item TTM\cite{ekambaram2024ttm}: Tiny Time Mixers, lightweight pre-trained models based on the TSMixer architecture, optimized for efficient transfer learning and low-resource environments.
\end{itemize}

By comparing UniCast with the original zero-shot versions of its backbones (Chronos, Timer, and TimesFM), we can isolate and quantify the contributions of multimodal cues and our PEFT-based adaptation. Furthermore, comparisons with MOMENT, MOIRAI, and TTM provide a broader context of UniCast’s performance relative to the current SOTA in time-series foundation modeling.

\subsection{Implementation Details}
Our UniCast integrates pretrained models in three modalities: vision, text, and time series. For the vision encoder, we utilize two state-of-the-art models: CLIP\cite{radford2021clip} and BLIP\cite{li2022blip}. For the text encoder, we employ Qwen2\cite{team2024qwen2} and LLaMA\cite{touvron2023llama}. For our TSFMs, we adopt Chronos\cite{ansari2024chronos}, Timer\cite{liu2024timer} and TimesFM\cite{das2024timesfm} due to their strong representational capabilities and compatibility with our prompting strategy.
All pretrained models were obtained from the Hugging Face Model Hub: \textbf{1) Vision Encoders}: \textit{openai/clip-vit-base-patch32}, \textit{Salesforce/blip-image-captioning-base}
\textbf{2) Text Encoders}: \textit{Qwen/Qwen2-1.5B-Instruct}, \textit{huggyllama/llama-7b}
\textbf{3) Time Series Models}: \textit{amazon/chronos-bolt-base}, \textit{thuml/timer-base-84m}, \textit{google/timesfm-2.5-200m-pytorch}.
To incorporate visual context, we transform time series into image representations using the approach proposed in ViTST \cite{li2023vitst}, implemented with the Python \textit{matplotlib} library.
For the text modality, we constructed a time-series description for each input window, conditioning the text encoder with \textit{Qwen/Qwen3-VL-8B-Instruct} as semantic context. Details can be found at Appendix~\ref{app:timeseriesdescription}.
All modules are optimized using the AdamW optimizer with a learning rate of $2\times10^{-5}$. Each model is trained for 10 epochs.
All experiments are implemented in PyTorch and executed on a single NVIDIA H100 NVL GPU with 94 GB memory.

\section{Result}
\subsection{Overall Performance}
To evaluate the effectiveness of UniCast, we compare it against a set of strong baselines, including Chronos, Timer, TimesFM, MOMENT, Moirai, and TTM.
For Chronos, Timer, and TimesFM, we additionally report \textbf{fully fine-tuned (FT)} results, which serve as strong upper-bound baselines.
All models are evaluated under identical training, validation, and test splits.
As shown in Table~\ref{tab:mainresult}, UniCast consistently outperforms the zero-shot (ZS) performance of all baselines and, notably, surpasses the FT variants of Chronos, Timer, and TimesFM on the majority of datasets.
This result is particularly significant given that \textbf{UniCast introduces only lightweight, parameter-efficient adaptations while keeping all backbone models frozen.}

The fact that UniCast outperforms fully fine-tuned backbones indicates that its gains cannot be attributed to increased model capacity or additional training, but instead arise from effective instance-conditioned multimodal control.
These improvements are observed consistently across diverse domains and data regimes, highlighting UniCast’s strong generalization ability and demonstrating that parameter-efficient multimodal adaptation can serve as a practical and scalable alternative to costly fine-tuning. All UniCast results reported in Table~\ref{tab:mainresult} correspond to the full model that incorporates Conditional Prompting (CP) and Modality Routing (MR).
To further analyze the contributions of individual design components in a controlled setting, we conduct ablation studies on three representative datasets selected from the full benchmark.
These datasets span diverse domains and data characteristics, allowing us to isolate the effects of conditional prompting and modality routing while keeping the experimental setup computationally tractable.
Unless specified, all ablation experiments use the same training protocol and hyperparameter settings as those in the main results.

\subsection{Effect of Pretrained Backbones}

\begin{table}[t]
    \caption[Effect of Pretrained Backbones]{Effect of Pretrained Time-Series Backbones. We compare zero-shot (ZS), full fine-tuning (FT), and UniCast built on top of three time-series foundation models: Chronos, Timer, and TimesFM. Across all backbones, UniCast consistently outperforms both ZS and FT variants, demonstrating that its performance gains are not tied to a specific time-series backbone but arise from the proposed parameter-efficient multimodal design. \textbf{Bold} indicates the best performance for each backbone-dataset pair.}
    \label{tab:pretrainedbackbones}
{\fontsize{9}{11}\selectfont
    \begin{tabular}{c|cc| ccc |c}
    \hline
    \textbf{TS} & \textbf{Vision} & \textbf{Text} & \textbf{NN5} & \textbf{Aus} & \textbf{Tour} & \textbf{Avg}\\
    \hline
    \multirow{4}{*}{\rotatebox{90}{\textbf{Chronos}}} & \multicolumn{2}{c|}{ZS} & 0.7913 &2.6701 &2.4682 &1.9765\\
    & \multicolumn{2}{c|}{FT} & 0.4635 &0.3606 &1.3896 &0.7379\\
    & CLIP & Qwen &0.4635 &0.3431 &1.3573 &0.7213\\
    & BLIP & LLaMA &\textbf{0.4577} &\textbf{0.3059} &\textbf{1.3467} &\textbf{0.7034}\\
    \hline
    \multirow{3}{*}{\rotatebox{90}{\textbf{Timer}}} & \multicolumn{2}{c|}{ZS} & 1.0986 & 1.4620$^*$ &2.6216 &1.7274\\
    & \multicolumn{2}{c|}{FT} & 0.9368 & 0.4839$^*$ &2.4492 &1.2900\\
    & CLIP & Qwen & \textbf{0.5087} & \textbf{0.2802}$^*$ & \textbf{1.3824} & \textbf{0.7238}\\
    \hline
    \multirow{3}{*}{\rotatebox{90}{\fontsize{7.5}{8}\selectfont \textbf{TimesFM}}} & \multicolumn{2}{c|}{ZS} & 1.7585 &2.2816 &3.1699 &2.4033\\
    & \multicolumn{2}{c|}{FT} & 0.5199 &0.5163 &1.5287 &0.8550\\
    &CLIP &Qwen &\textbf{0.4690} &\textbf{0.2904} &\textbf{1.3888} &\textbf{0.7161}\\
    \hline
    \end{tabular}
}
\begin{flushleft}
\footnotesize
TS: Time Series
\end{flushleft}
\vspace{-0.5cm}
\end{table}

To assess whether the effectiveness of UniCast depends on a specific time-series backbone, we conduct experiments using multiple pretrained TSFMs, including Chronos, Timer, and TimesFM, as the backbone of UniCast. For each backbone, we compare its zero-shot (ZS) and fully fine-tuned (FT) variants with UniCast built on top of the same model, ensuring a controlled and fair comparison. The results are summarized in Table~\ref{tab:pretrainedbackbones}.
Across all backbones and datasets, UniCast consistently outperforms both the ZS and FT variants of the corresponding backbone. Notably, UniCast often surpasses fully fine-tuned backbones despite introducing only lightweight, parameter-efficient adaptations while keeping all backbone parameters frozen. This indicates that UniCast's performance gains are not attributable to backbone-specific tuning but rather arise from its architecture-level multimodal control via conditional prompting and modality routing.
These results suggest that UniCast functions as a general, backbone-agnostic enhancement layer for time-series foundation models.
Rather than replacing existing TSFMs, UniCast complements them by enabling instance-conditioned multimodal adaptation that is more effective than fine-tuning alone.

\subsection{Impact of Conditional Prompting and Modality Routing}

\begin{figure}
    \centering
    \includegraphics[width=\columnwidth]{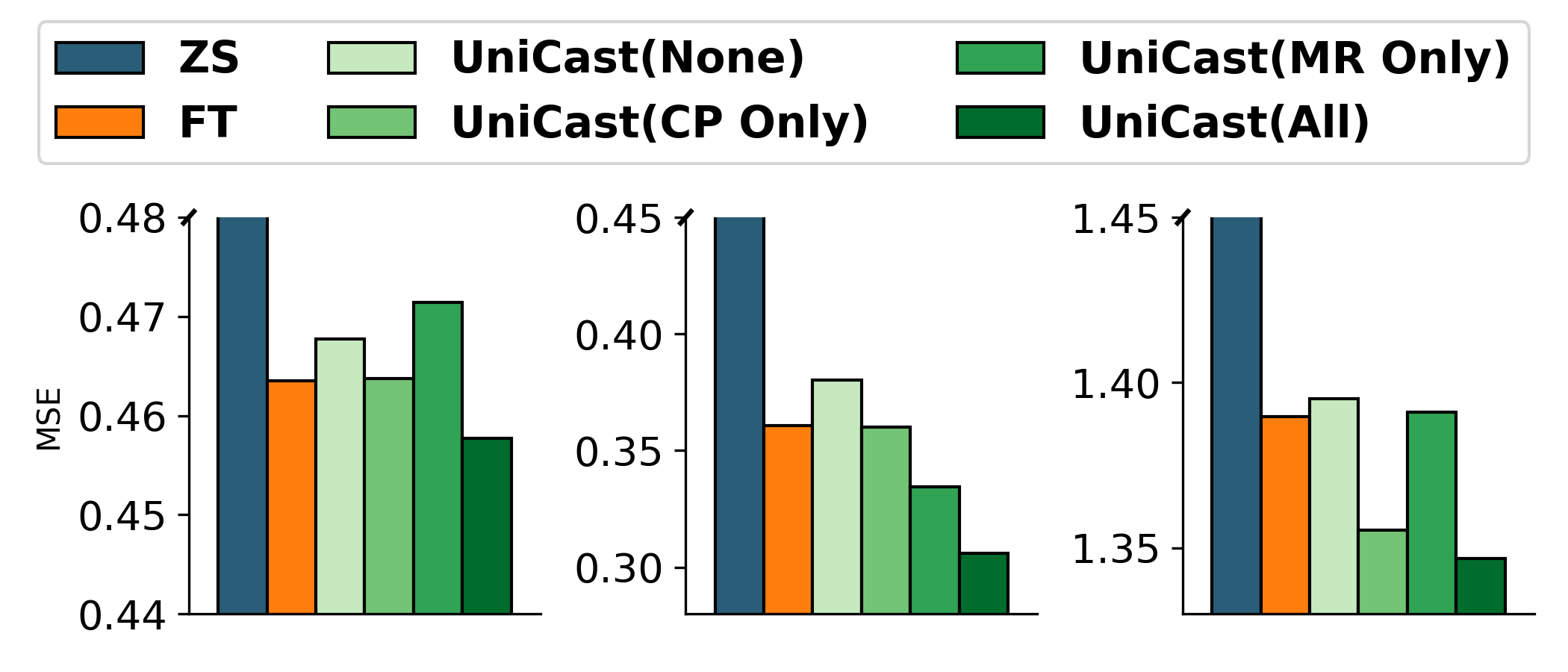}
    \makebox[\columnwidth][c]{%
        \hspace{0.05\columnwidth}\textbf{(a) NN5 Daily}%
        \hspace{0.12\columnwidth}\textbf{(b) Aus Elec}%
        \hspace{0.11\columnwidth}\textbf{(c) Tourism}%
    }
    \caption[Impact of Conditional Prompting and Modality Routing.]{Ablation study on Conditional Prompting and Modality Routing. We compare zero-shot (ZS), full fine-tuning (FT), and UniCast variants with different design components: without CP and MR (None), with CP only, with MR only, and with both components enabled (All). Both CP and MR individually improve performance over static or heuristic baselines, while combining both components consistently yields the lowest forecasting error, highlighting their complementary roles in instance-conditioned multimodal adaptation. Exact values can be found at Table~\ref{tab:cpmrablation} in Appendix~\ref{app:cpmrablation}.}

    \label{fig:cpmrablation}
    \vspace{-0.2cm}
\end{figure}

In this subsection, we examine the impact of Conditional Prompting (CP) and Modality Routing (MR) on UniCast performance.
To evaluate whether CP and MR provide benefits beyond static or heuristic strategies,  we replace CP with static soft prompting and MR with heuristic modality fusion, respectively.
Figure~\ref{fig:cpmrablation} shows that applying either CP or MR leads to performance improvements,  while combining both components consistently yields the best results.
These results indicate that CP and MR contribute complementary functionalities. Conditional Prompting enables instance-conditioned adaptation by providing contextual prompts,  while Modality Routing selectively injects modality-specific information based on the evolving temporal representations. When used together, UniCast can effectively leverage auxiliary modalities in a controlled and adaptive manner.
Overall, this ablation study highlights the importance of explicitly modeling instance-conditioned prompting and selective modality utilization, rather than relying on static prompts or heuristic fusion schemes.
Exact numerical results for each design choice are reported in Table~\ref{tab:cpmrablation} in Appendix~\ref{app:cpmrablation}.
We further analyze implementation variants of CP and MR in Appendix~\ref{app:cpmrimplementation}.

\begin{figure*}[t]
\centering
\begin{subfigure}{0.24\linewidth}
    \centering
    \includegraphics[width=\linewidth]{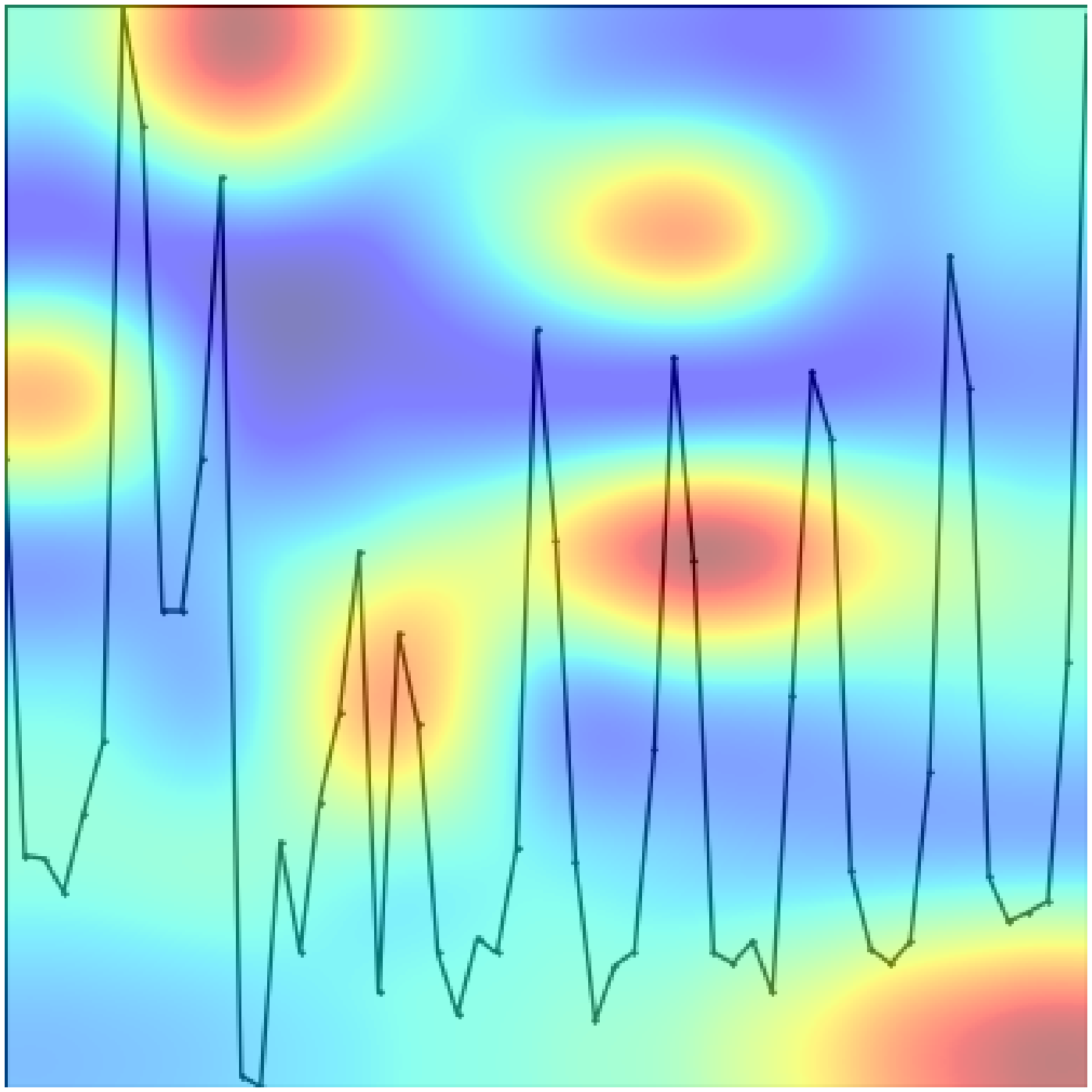}
    \caption{Layer 1 to 3}
\end{subfigure}
\begin{subfigure}{0.24\linewidth}
    \centering
    \includegraphics[width=\linewidth]{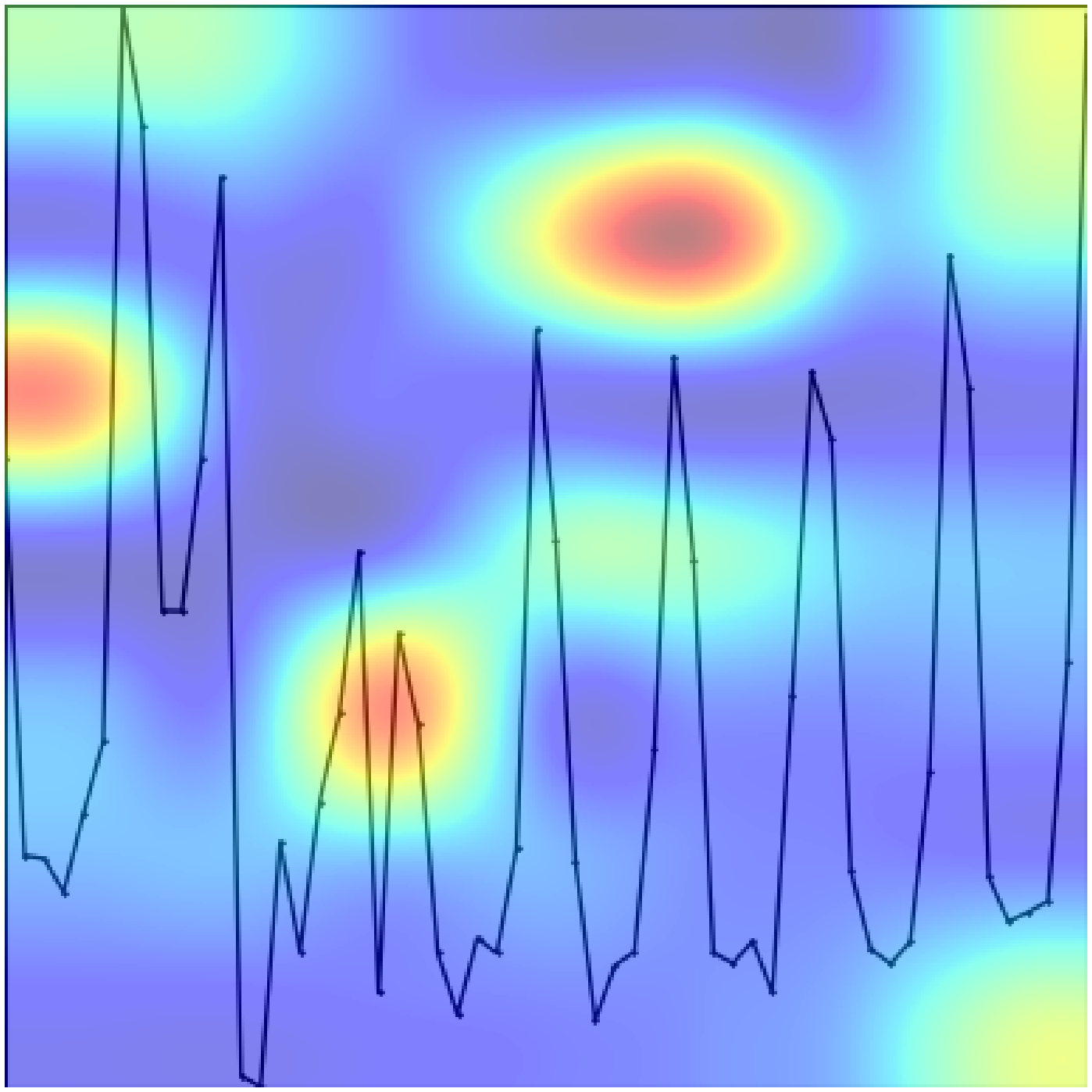}
    \caption{Layer 4 to 6}
\end{subfigure}
\begin{subfigure}{0.24\linewidth}
    \centering
    \includegraphics[width=\linewidth]{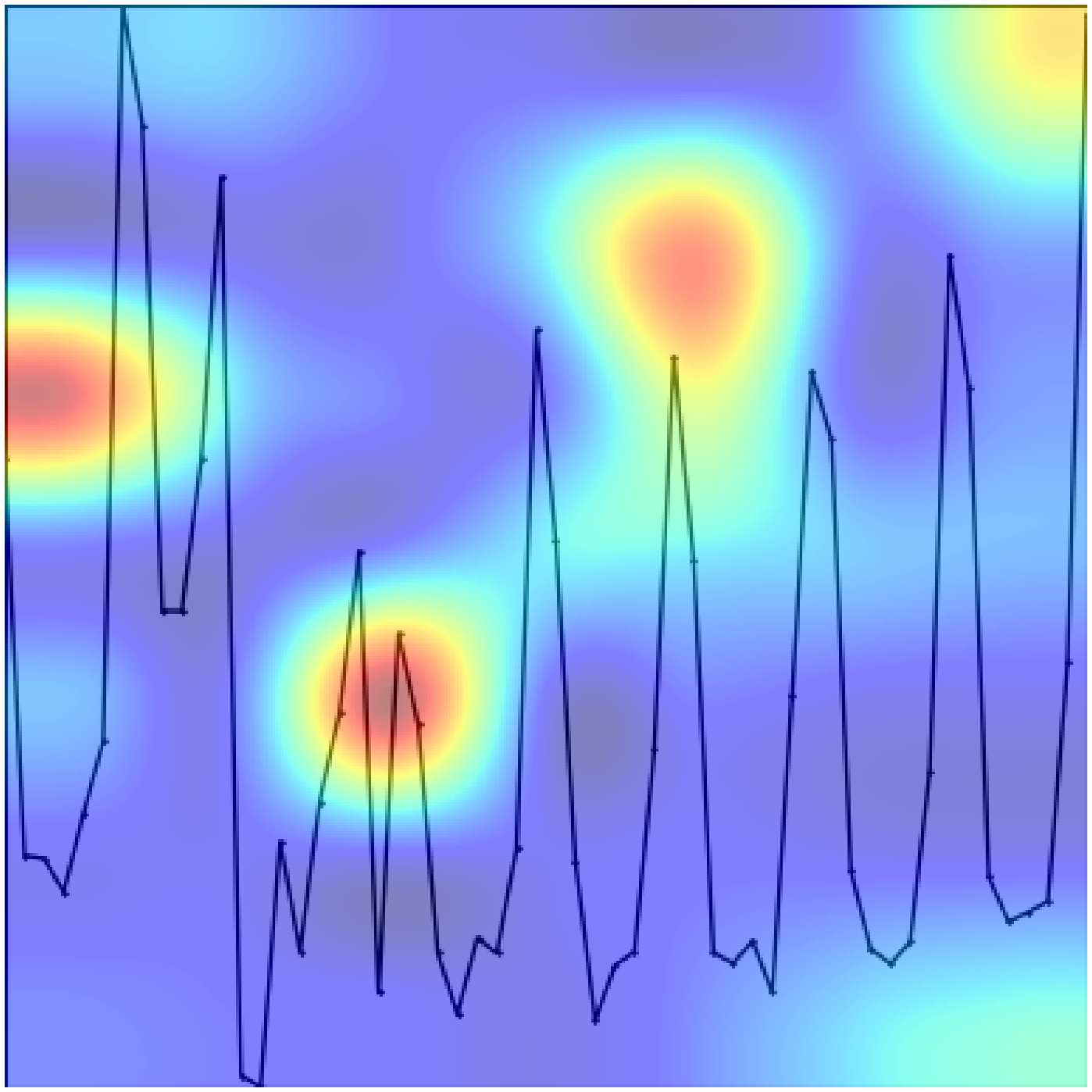}
    \caption{Layer 7 to 9}
\end{subfigure}
\begin{subfigure}{0.24\linewidth}
    \centering
    \includegraphics[width=\linewidth]{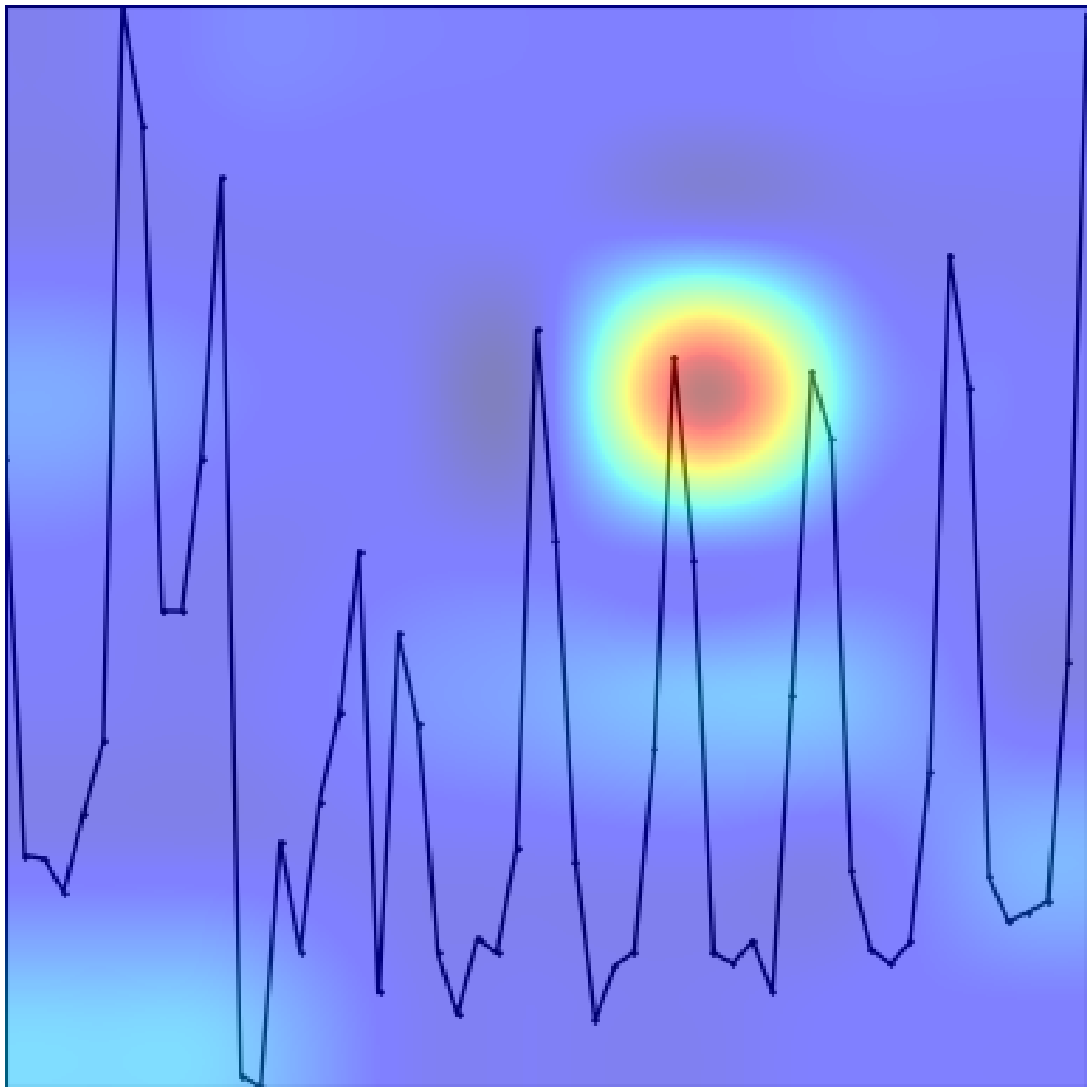}
    \caption{Layer 10 to 12}
\end{subfigure}
\caption[Attention Heatmaps of BLIP.]{Attention heatmaps of the BLIP vision backbone. Each subfigure visualizes attention patterns at progressively deeper layers, from early (Layer 1 to 3) to later (Layer 10 to 12) stages of processing, illustrating how visual relevance evolves during prediction.}
\label{fig:blipheatmap}
\end{figure*}

\subsection{Effect of Vision and Text Modalities}

\begin{table}[t]
    \caption[Modality Ablations]{Effect of vision and text modalities. We evaluate UniCast under different modality configurations, including vision-only, text-only, and combined vision–text settings, as well as different pretrained vision (CLIP, BLIP) and text (Qwen, LLaMA) backbones. ZS and FT denote zero-shot and full fine-tuning of the time-series backbone (Chronos), respectively, and PT denotes prompt tuning of the same backbone without using multimodality. Using both vision and text consistently improves performance over single-modality variants, indicating their complementary effects. \textbf{Bold} indicates the best performance and \underline{underlined} the second best for each dataset.}
    \label{tab:modalityablation}
{\fontsize{9}{11}\selectfont
    \begin{tabular}{cc| ccc |c}
    \hline
    \textbf{Vision} & \textbf{Text} & \textbf{NN5} & \textbf{Aus-Elec} & \textbf{Tourism} & \textbf{Avg}\\
    \hline
    \multicolumn{2}{c|}{ZS} & 0.7913 & 2.6701 & 2.4682 & 1.9765\\
    \multicolumn{2}{c|}{FT} & 0.4635 & 0.3606 & 1.3896 & 0.7379\\
    \multicolumn{2}{c|}{PT} & 0.4890 & 0.5613 & 1.4738 & 0.8414\\\hline
    \multicolumn{2}{c|}{\textbf{UniCast(Ours)}} &&&&\\
    CLIP & $\times$ & 0.4658 & 0.3450 & 1.3535 & 0.7214\\
    BLIP & $\times$ & 0.4625 & 0.3517 & 1.3559 & 0.7234\\
    $\times$ & Qwen & 0.6680 & 0.3753 & 1.4060 & 0.8164\\
    $\times$ & LLaMA & 0.4706 & 0.3823 & 1.3680 & 0.7403\\
    CLIP & Qwen & 0.4635 & 0.3431 & 1.3573 & 0.7213\\
    BLIP & Qwen & 0.4601 & 0.3320 & \underline{1.3499} & 0.7140\\
    CLIP & LLaMA & \textbf{0.4558} & \underline{0.3104} & 1.3534 & \underline{0.7065}\\
    BLIP & LLaMA & \underline{0.4577} & \textbf{0.3059} & \textbf{1.3467} & \textbf{0.7034}\\
    \hline
    \end{tabular}
}
\end{table}

We conduct ablation studies to analyze (1) the individual and combined effects of vision and text modalities and (2) the sensitivity of UniCast to different pretrained vision and text backbones. The results are summarized in Table~\ref{tab:modalityablation}.
Compared with zero-shot, full fine-tuning, prompt tuning baselines (with only time-series), using either vision or text alone already yields substantial performance improvements, often matching or exceeding those of fine-tuned time-series models. 
When both vision and text modalities are used together, UniCast consistently outperforms all single-modality variants and further improves upon full fine-tuning. This demonstrates that vision and text provide complementary signals that are selectively exploited in an instance-conditioned manner, rather than being indiscriminately fused.
We further evaluate four backbone configurations using two vision models (CLIP, BLIP) and two text models (Qwen, LLaMA). Across all backbone combinations, UniCast exhibits stable and consistent performance gains. These results indicate that UniCast does not rely on a specific pretrained vision or text model, and that its improvements primarily stem from the proposed multimodal prompting and routing design, rather than from a particular backbone choice.

\begin{figure*}[t]
\centering

\begin{subfigure}{0.19\linewidth}
    \centering
    \includegraphics[width=\linewidth]{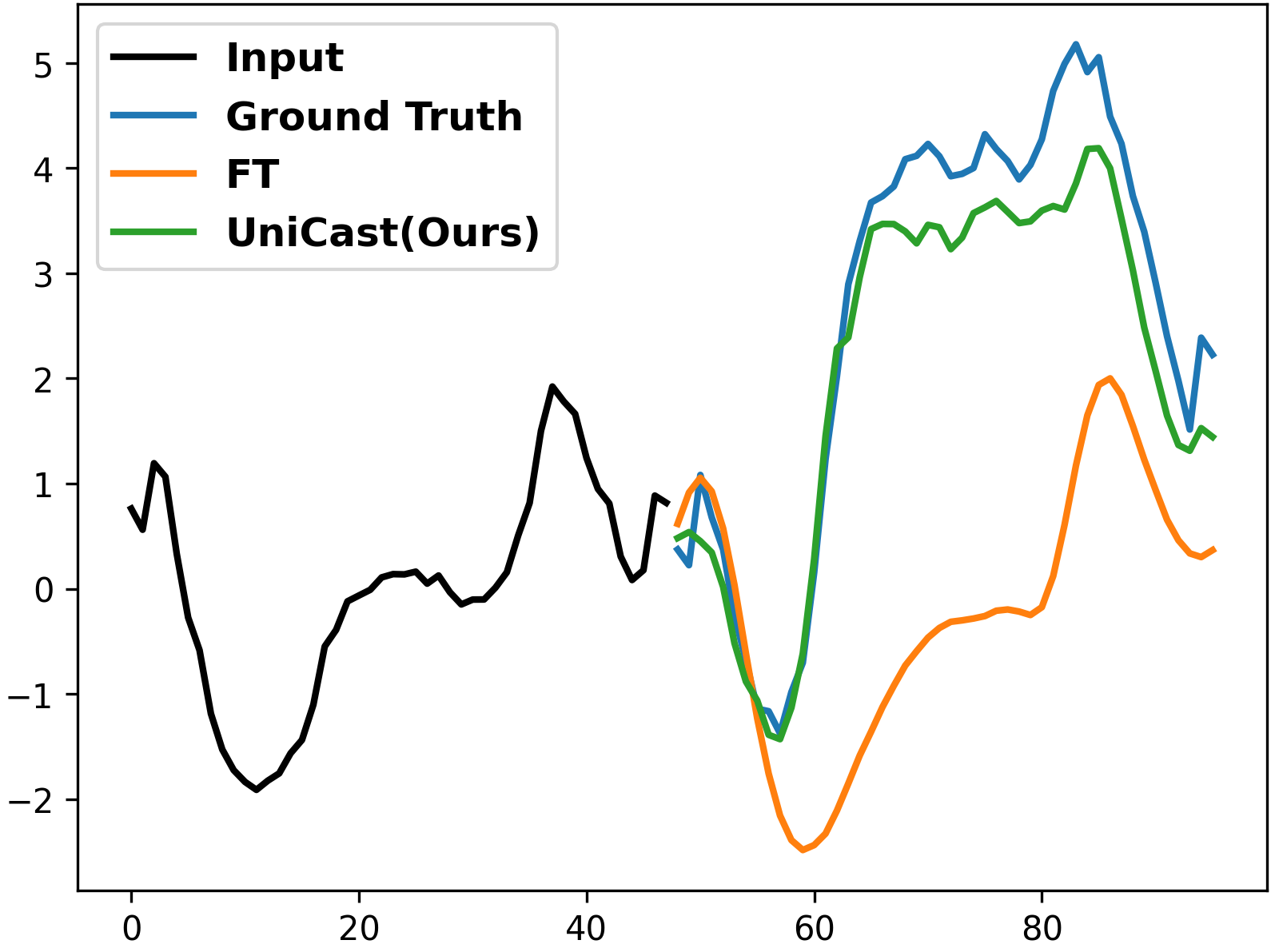}
    \caption{Aus-Elec}
\end{subfigure}
\hfill
\begin{subfigure}{0.19\linewidth}
    \centering
    \includegraphics[width=\linewidth]{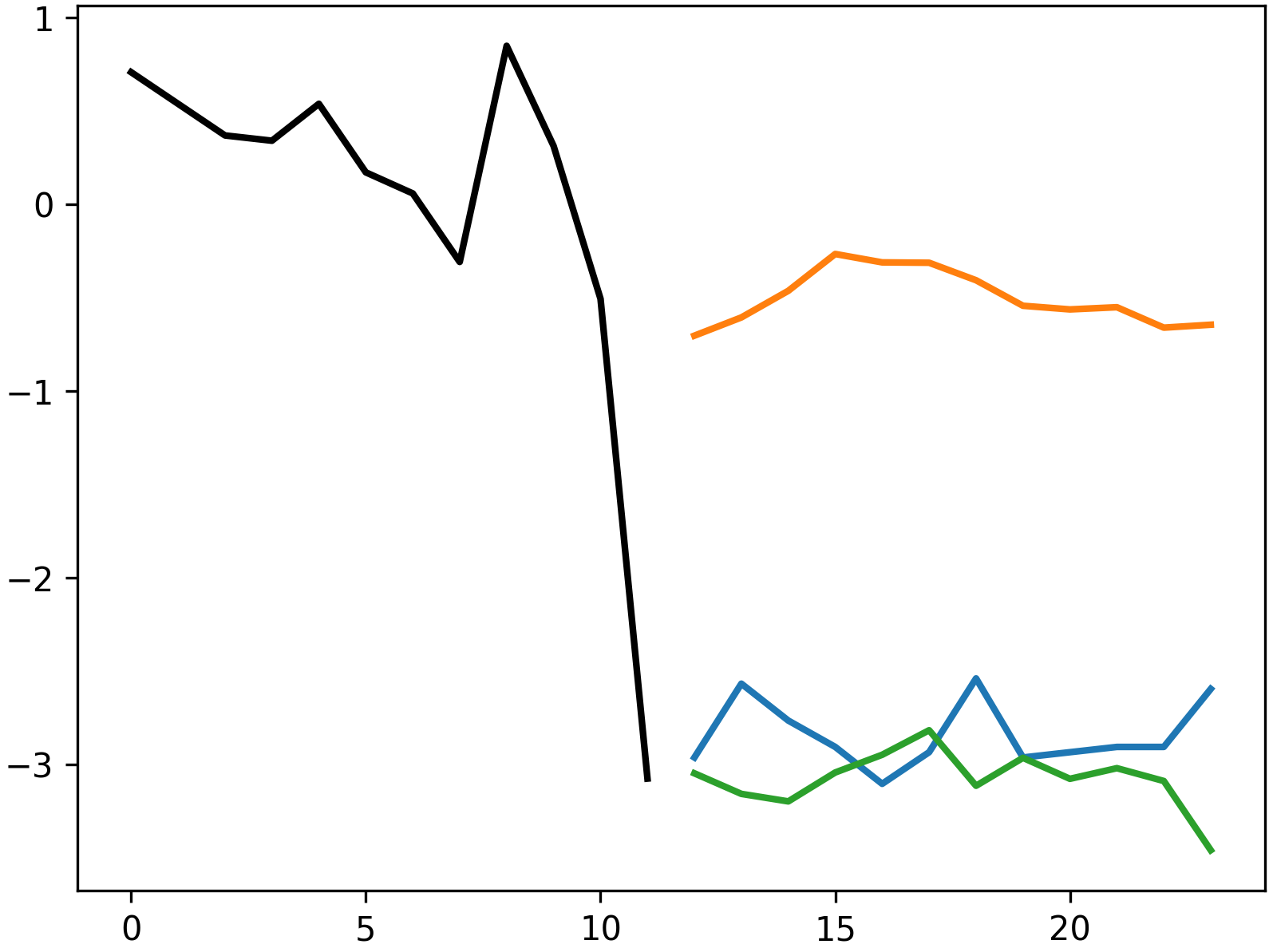}
    \caption{Hospital}
\end{subfigure}
\hfill
\begin{subfigure}{0.19\linewidth}
    \centering
    \includegraphics[width=\linewidth]{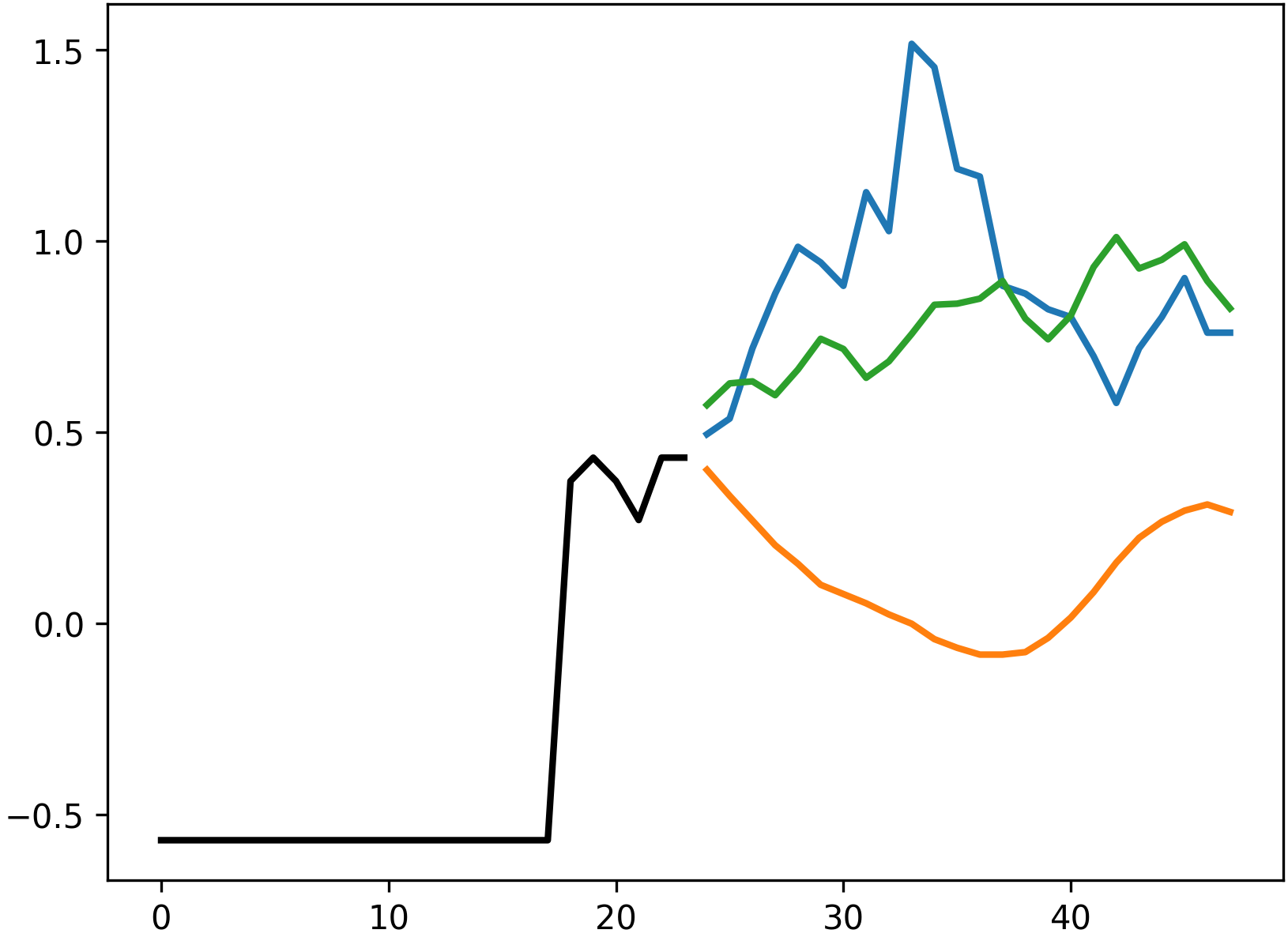}
    \caption{ETTh1}
\end{subfigure}
\hfill
\begin{subfigure}{0.19\linewidth}
    \centering
    \includegraphics[width=\linewidth]{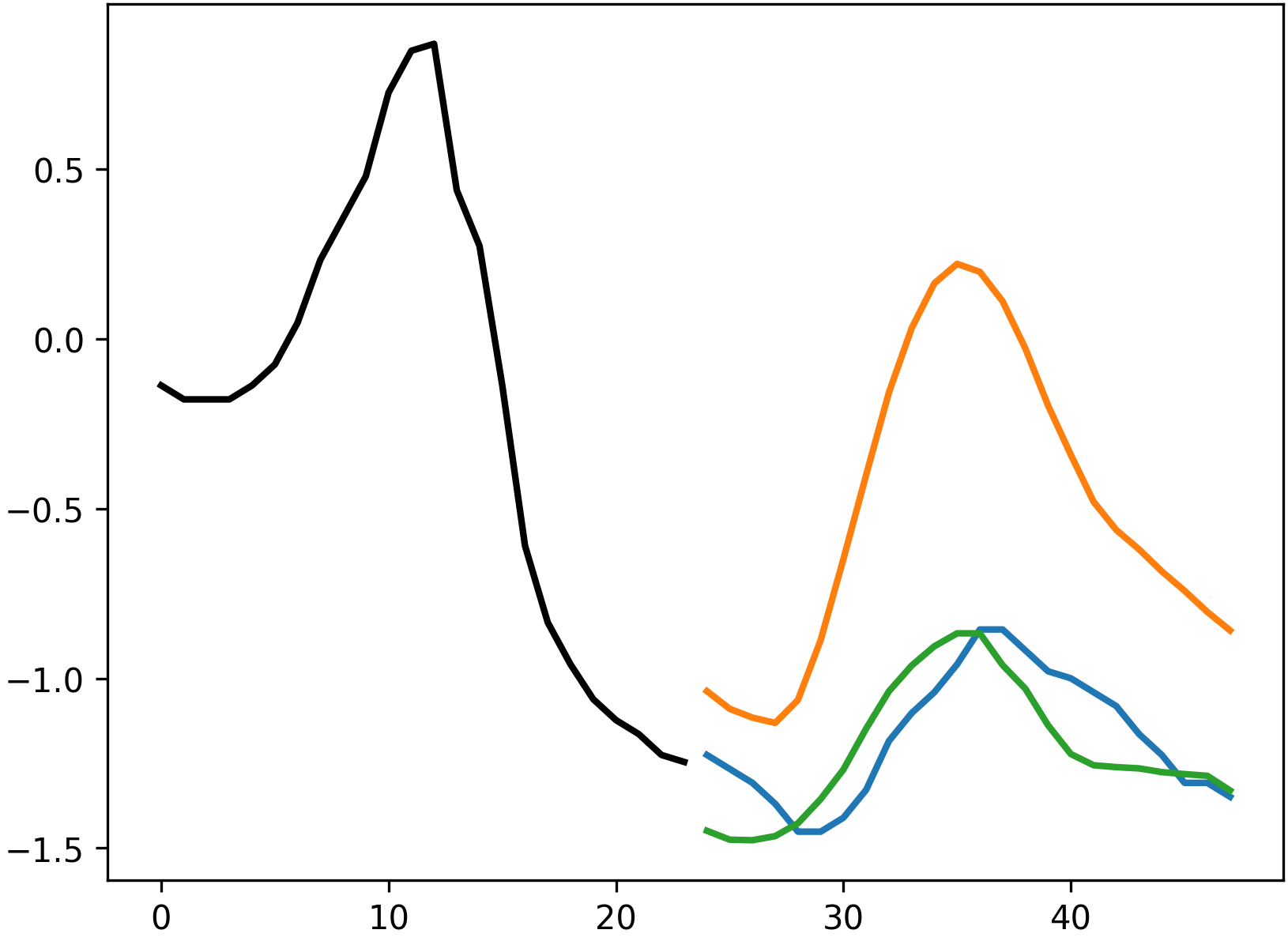}
    \caption{ETTh2}
\end{subfigure}
\hfill
\begin{subfigure}{0.19\linewidth}
    \centering
    \includegraphics[width=\linewidth]{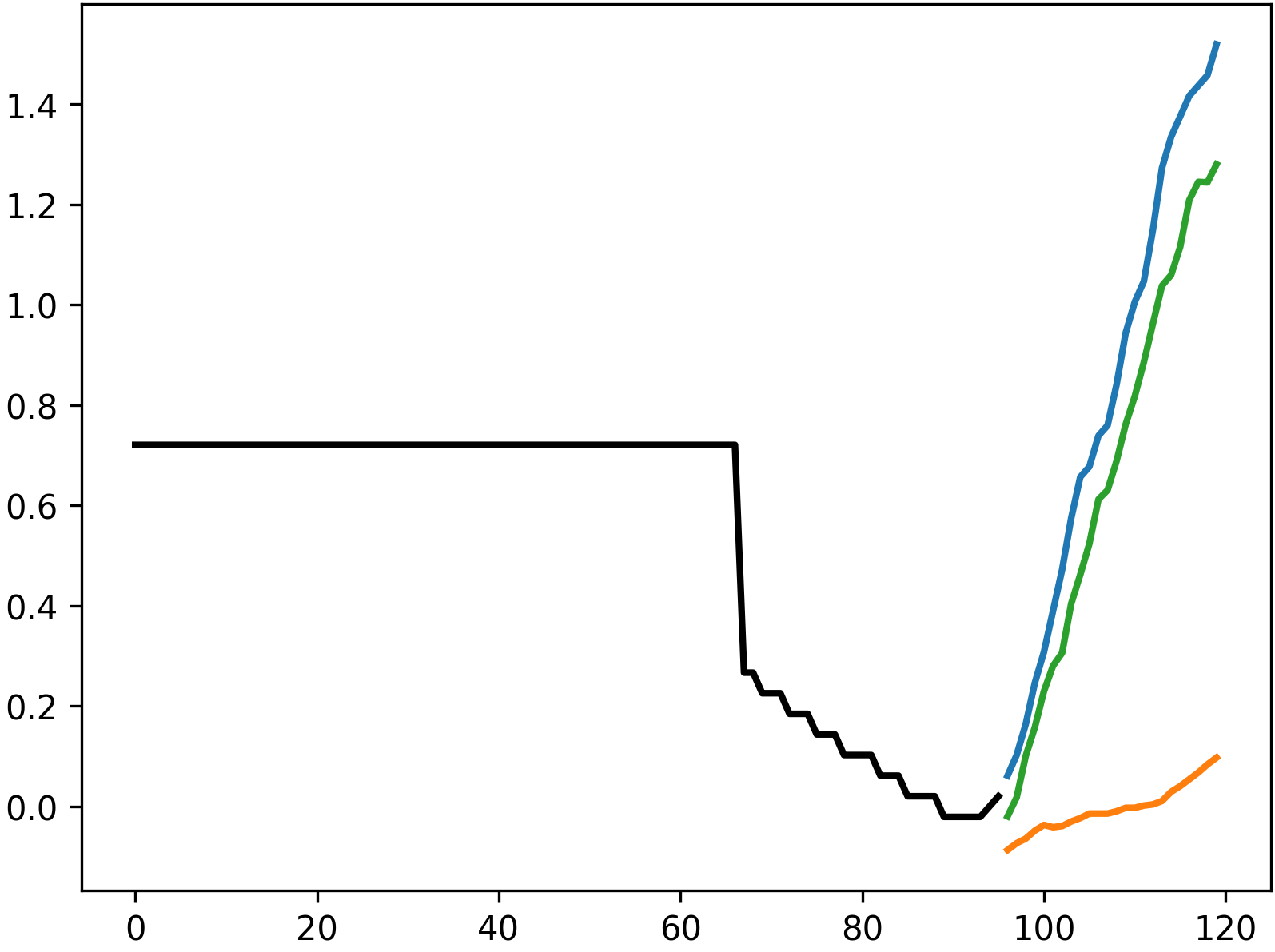}
    \caption{ETTm2}
\end{subfigure}

\caption[Forecasting Examples]{Forecasting examples. Subfigures show results on five datasets: (a) Aus-Elec, (b) Hospital, (c) ETTh1, (d) ETTh2 and (e) ETTm2. Each subfigure compares the forecasts produced by a fully fine-tuned time-series model \textcolor{orange}{(FT)} and \textcolor{ForestGreen}{UniCast(Ours)} against the \textcolor{RoyalBlue}{Ground Truth}. The black curve denotes the input context, while colored curves correspond to predictions over the forecast horizon.}
\label{fig:forecastingexamples}
\end{figure*}

\section{Qualitative Analysis}
Beyond quantitative performance, we provide qualitative analyses to better understand how UniCast leverages multimodal information for time-series forecasting.
Specifically, we visualize attention heatmaps across modalities and present forecasting examples to illustrate how multimodal signals contribute to prediction behavior.

\subsection{Attention Visualization}
To examine how UniCast performs instance-conditioned multimodal utilization during prediction, we visualize layer-wise attention heatmaps for the vision, text, and time-series modalities.
These visualizations provide insight into how modality relevance evolves across layers as temporal representations are progressively refined. Figure~\ref{fig:blipheatmap} presents a representative example using the BLIP vision backbone.
In the early layers, attention is broadly distributed across the visual input, indicating that UniCast initially captures coarse, global visual context without strong preference for specific regions. As processing proceeds to deeper layers, attention gradually concentrates on localized regions that are more predictive of the target time-series behavior. This transition from diffuse global attention to focused, task-relevant regions reflects UniCast’s ability to filter auxiliary information and selectively emphasize salient visual cues for forecasting.
Similar qualitative trends are observed for the text and time-series modalities. For text inputs, attention increasingly focuses on semantically meaningful tokens related to trends or events, while less informative tokens receive reduced weight. For time-series inputs, attention becomes concentrated on temporally salient segments that are most relevant for future prediction. These observations indicate that UniCast selectively exploits complementary multimodal signals in an instance-conditioned manner, rather than indiscriminately fusing all available modalities.
Due to space limitations, additional attention visualizations, including results using the CLIP vision backbone and further examples for text and time-series modalities, are provided in Appendix~\ref{app:heatmaps}.

\subsection{Forecasting Examples}

Figure~\ref{fig:forecastingexamples} presents representative forecasting examples comparing UniCast with fully fine-tuned time-series models under distribution shifts between the input context and the prediction horizon.
In these scenarios, the statistical characteristics of the future target segment differ substantially from those observed in the input window, posing challenges for models that primarily rely on recent temporal patterns.
Fully fine-tuned time-series models tend to extrapolate historical trends from the input context, leading to noticeable deviations when the underlying dynamics change.

In contrast, UniCast produces forecasts that more closely align with the ground-truth trajectories, particularly around turning points, abrupt trend changes, and amplitude variations.
These examples indicate that multimodal contextual information enables UniCast to better adapt to distribution shifts by providing auxiliary cues beyond the raw time-series history.
As a result, UniCast exhibits improved robustness in challenging forecasting regimes where temporal continuity alone is insufficient.
This qualitative evidence complements the quantitative results, demonstrating that UniCast improves not only average forecasting accuracy but also predictive reliability under non-stationary conditions.

\section{Parameter Efficiency}
Table~\ref{tab:parameterefficiency} reports a detailed analysis of the parameter efficiency of UniCast across different configurations, including the total number of parameters, trainable parameters, and their relative proportions.
We compare the full UniCast model with several ablated variants to understand how trainable capacity is allocated across components.
Across all main configurations, UniCast operates in a strongly parameter-efficient regime. Despite incorporating large pretrained vision and text encoders, only approximately 5--6\% of the total parameters are trainable. All pretrained backbones, including the time-series model and multimodal encoders, remain frozen, and learning is confined to lightweight prompt generators, routing modules, and projection layers.
When measured relative to the Chronos backbone, UniCast updates between 56\% (Qwen-based variants) and 193\% (LLaMA-based variants) of Chronos’s parameter count.
This increases reflects the addition of small trainable adaptation modules rather than modifications to the backbone itself, while enabling UniCast to leverage contextual information from much larger frozen multimodal models.
The ablation results further clarify the source of trainable parameters.
Conditional Prompting (CP) accounts for most of the trainable capacity by employing Transformer-based modules to model instance-specific multimodal dependencies.
In contrast, static soft prompting introduces orders of magnitude fewer parameters, but at the cost of reduced adaptability.
This trade-off highlights that the additional parameters introduced by CP are essential for enabling input-conditioned multimodal interactions, which are central to UniCast’s design.
Overall, the parameter analysis demonstrates that UniCast allocates its limited trainable capacity primarily toward modeling multimodal context and conditional control, while keeping the number of updated parameters small relative to the full model size.
This design allows UniCast to achieve strong performance gains without sacrificing the scalability and generalization benefits of pretrained foundation models.

\section{Conclusion}
We introduced UniCast, a parameter-efficient multimodal framework for time-series forecasting that enables instance-level control over auxiliary modality usage. By separating modality-aware context extraction from modality utilization, UniCast selectively integrates vision and text information conditioned on the current temporal state, while keeping all pretrained models frozen.
Empirical results show that UniCast consistently outperforms strong zero-shot and fully fine-tuned TSFM baselines across diverse datasets, with notable gains under distribution shift. Our findings suggest that forecasting performance can be substantially improved through adaptive multimodal control rather than increased model size or extensive fine-tuning.
Beyond accuracy improvements, UniCast exhibits enhanced robustness under non-stationary conditions and provides interpretable signals of modality relevance through its routing mechanism. We believe UniCast offers a scalable and practical foundation for deploying multimodal time-series forecasting systems in real-world applications where contextual information is heterogeneous and instance-dependent.


\bibliography{unicast.bbl}

\begin{table*}[t]
  \centering
  \caption[Summary statistics of Dataset]{Summary statistics of Dataset.}
  \label{tab:datasetstatistics}
{\fontsize{9}{11}\selectfont
  \begin{tabular}{ccccccccc}
    \toprule
    \multirow{2}{*}{\textbf{Dataset}} & \multirow{2}{*}{\textbf{Domain}}
     & \multirow{2}{*}{\textbf{Frequency}} & \multirow{2}{*}{\textbf{\# of Series}} & \multicolumn{3}{c}{\textbf{Series Length}} & \multirow{2}{*}{$C$} & \multirow{2}{*}{$H$}\\
     \cmidrule{5-7}
     & & & & \textbf{min} & \textbf{avg} & \textbf{max} & &\\
    \midrule
    NN5 Daily & Finance (cash withdrawal volumes) & 1 Day & 111 & - & 791 & - & 56 & 56 \\
    Australian Electricity & Energy (electricity demand) & 30 Minutes & 5 & 230736 & 231052 & 232272 & 48 & 48 \\
    Tourism Monthly & Others (tourism demand) & 1 Month & 366 & 91 & 298 & 333 & 24 & 24 \\
    COVID-19 Deaths & Health (death across countries) & 1 Day & 266 & - & 212 & - & 30 & 30\\
    Car Parts & Retail (car component sales) & 1 Month & 2674 & - & 51 & - & 12 & 12 \\
    CIF 2016 & Banking (real-time banking demand) & 1 Month & 72 & 28 & 98 & 120 & 12 & 12 \\
    Dominick & Retail (stock-keeping units) & 1 Week & 100014 & 201 & 296 & 399 & 8 & 8 \\
    Hospital & Health (patient count) & 1 Month & 767 & - & 84 & - & 12 & 12 \\
    ETTh & Energy (transformer temperature) & 1 Hour & 14 & - & 17420 & - & 24 & 24 \\
    ETTm & Energy (transformer temperature) & 15 Minutes & 14 & - & 69680 & - & 96 & 24 \\
    \bottomrule
  \end{tabular}
}
\begin{flushright}
\footnotesize
$C$ Context Length. $H$ Forecast Length.\\
\end{flushright}
\end{table*}
\newpage
\appendix

\section{Dataset Details}
\label{app:datasetdetail}
To evaluate the effectiveness of our \textbf{UniCast}, we adopt a subset of benchmarks used in the zero-shot evaluation of Chronos \cite{ansari2024chronos}. For a fair comparison, we follow the same settings as Chronos, using identical context lengths $C$ and forecast horizons $H$. Summary statistics including forecasting configuration for all datasets are presented in Table~\ref{tab:datasetstatistics}. 
Each time series is split into training, validation, and test sets using a 60:20:20 ratio based on its sequence length. In general, we apply standardization within each input-target window pair. However, due to dataset-specific properties, such as short sequence lengths or constant segments, we adopt alternative preprocessing strategies where necessary, as detailed below.
\noindent\textbf{\textit{1) COVID-19 Deaths}}\footnotemark[1] contains 266 daily time series representing COVID-19 death counts across various countries and states, from 22/01/2020 to 20/08/2020. Each series consists of only 212 time steps, which is relatively short given our context and forecast window sizes. To ensure proper sample diversity, we shuffle the time series before applying the 6:2:2 split. Since many context windows contain constant values, standardization is applied over the entire series to avoid division errors.
\textbf{\textit{2) NN5 Daily}}\footnotemark[2] includes 111 daily time series from the UK banking sector, capturing ATM cash withdrawal volumes. We adopt the preprocessing strategy mentioned earlier.
\textbf{\textit{3) Car Parts}}\footnotemark[1] comprises monthly sales data for various car components between January 1998 and March 2002. Due to the limited sequence lengths, we apply series-level shuffling followed by a 6:2:2 split. Additionally, to prevent numerical instability caused by constant windows, standardization is performed over entire sequences.
\textbf{\textit{4) Australian Electricity}}\footnotemark[1] consists of half-hourly electricity demand data from five Australian states. Each series contains over 200,000 time steps. To reduce computational overhead while preserving temporal diversity, we use only the first 15,000 steps per series for training and evaluation.
\textbf{\textit{5) CIF 2016}}\footnotemark[1] originates from the CIF 2016 forecasting competition and includes 72 monthly banking-related time series: 24 real and 48 synthetic. Due to their short lengths, we apply the same shuffle-and-split strategy and use series-level standardization.
\textbf{\textit{6) Dominick}}\footnotemark[1] contains over 100,000 weekly time series measuring the profitability of individual stock-keeping units in a retail setting. For computational feasibility, we randomly sample 100 series after shuffling and apply standard preprocessing.
\textbf{\textit{7) Hospital}}\footnotemark[1] includes monthly time series reflecting patient counts related to medical products, recorded from January 2000 to December 2006. Due to the limited sequence lengths, we follow the same preprocessing approach as with Car Parts and CIF 2016.
\textbf{\textit{8) Tourism Monthly}}\footnotemark[1] is derived from the Kaggle Tourism Forecasting competition and includes a collection of short monthly tourism-related time series. As with similar datasets, we shuffle and split the series and apply standardization at the full-series level.
\textbf{\textit{9) ETT}}\footnotemark[3] contains oil temperature measurements and six types of external power load features collected from two electricity transformers in China over a two-year period. The dataset has four variants: ETTh1, ETTh2, ETTm1, and ETTm2, corresponding to hourly (`h`) and 15-minute (`m`) sampling frequencies. In our experiments, we use only the `OT` series, which represents the target oil temperature.

\footnotetext[1]{We obtained those datasets from https://huggingface.co/datasets/autogluon/chronos\_datasets}
\footnotetext[2]{We obtained NN5 dataset from https://zenodo.org/records/4656117 due NN5 dataset in HuggingFace Chronos dataset repository has missing values.}
\footnotetext[3]{We obtained ETT dataset from https://github.com/zhouhaoyi/ETDataset.}


\section{Time Series Description}
\label{app:timeseriesdescription}

To provide textual information corresponding to the input time series to the text encoder, we generate a natural-language description for each input window.
Specifically, we leverage the \textit{Qwen/Qwen3-VL-8B-Instruct} via Huggingface to produce a \emph{time series description} that summarizes the temporal characteristics of the input.

For each input window, the raw time series values are provided to the model along with a fixed prompt that instructs the model to analyze the data and describe its temporal structure.
The prompt explicitly asks the model to focus on two key aspects: (i) the overall trend, including its direction, shape, and consistency, and (ii) the presence of seasonality or repeating patterns.
If the evidence for either trend or seasonality is weak or insufficient, the model is instructed to explicitly state this uncertainty.

The prompt used to generate the time series descriptions is shown below:

\begin{quote}
\small
\texttt{You are given a sequence of time series values.\\
\\
The time series data is:\\
\{time series\}\\
\\
Analyze the time series and focus primarily on identifying:\\
- the overall trend (direction, shape, and consistency)\\
- any evidence of seasonality or repeating patterns\\
\\
Return your answer strictly in valid JSON format and do not include any extra text.\\
\\
The JSON must contain only one field:\\
- "description": a concise and clear natural-language description of the trend and seasonality observed in the time series. If seasonality is weak or not detectable, explicitly state that.\\
\\
If the data is insufficient to confidently determine trend or seasonality, clearly express this uncertainty in the description.}
\end{quote}

The generated description is then used as the textual input corresponding to the same time series window.
This approach allows the text encoder to access high-level semantic information about the temporal dynamics of the input without relying solely on raw numerical values.

\section{UniCast Computation Flow}
\label{app:algorithm}

\begin{algorithm}[t]
\caption{UniCast Computation Flow}
\label{alg:algorithm}
\begin{algorithmic}[1]

\Require
Time-series input $X \in \mathbb{R}^{T \times D}$, 
visual input $V$, textual input $L$

\Statex
\State \textbf{Tokenization and Patch Embedding}
\State $O_v = \mathrm{Tokenize}_v(V)$
\State $O_t = \mathrm{Tokenize}_t(L)$
\State $O_{ts} = \mathrm{Patch}(X)$

\Statex
\State \textbf{Modality-wise Pooling}
\State $\bar{O}_m = \mathrm{MeanPool}(O_m)$, \quad $m \in \{v,t,ts\}$

\Statex
\State \textbf{Modality-aware Context Distillation}
\State $c_m = \mathrm{ContextDistiller}_m(\bar{O}_m)$, \quad $m \in \{v,t,ts\}$

\Statex
\State \textbf{Vision and Text Encoding with Conditional Prompts}
\State $p_v = \mathrm{PromptGen}_v([c_v;\bar{O}_v])$
\State $E_v = \mathrm{VisionEnc}([p_v; O_v])$

\State $p_t = \mathrm{PromptGen}_t([c_t;\bar{O}_t])$
\State $E_t = \mathrm{TextEnc}([p_t; O_t])$

\Statex
\State \textbf{Projection to Time-Series Space}
\State $E_v^{ts} = \mathrm{Proj}_{v\rightarrow ts}(E_v)$
\State $E_t^{ts} = \mathrm{Proj}_{t\rightarrow ts}(E_t)$
\State $S = [E_v^{ts}; E_t^{ts}]$

\Statex
\State \textbf{Conditional Time-Series Prompt Generation}
\State $\bar{E}_v^{ts} = \mathrm{MeanPool}(E_v^{ts})$
\State $\bar{E}_t^{ts} = \mathrm{MeanPool}(E_t^{ts})$
\State $p_{ts} = \mathrm{PromptGen}_{ts}([\bar{E}_v^{ts}; \bar{E}_t^{ts}; \bar{O}_{ts}])$

\Statex
\State \textbf{Modality Routing Initialization}
\State $O_{ts}^{0} = O_{ts} + \mathrm{CrossAttn}(Q=O_{ts}, K=S, V=S)$
\State $H^{0} = [p_{ts}; O_{ts}^{0}]$

\Statex
\State \textbf{Iterative Modality Routing and TSFM Processing}
\For{$k = 0,1,\dots,L-1$}
    \State $H^{k+\frac{1}{2}} = \mathrm{TSFMBlock}_{k+1}(H^{k})$
    \State $H^{k+1} = H^{k+\frac{1}{2}} + \mathrm{CrossAttn}(Q=H^{k+\frac{1}{2}}, K=S, V=S)$
\EndFor

\Statex
\State \textbf{Forecasting}
\State $\hat{y} = f_{\text{head}}(H^{L})$

\end{algorithmic}
\end{algorithm}

Algorithm~\ref{alg:algorithm} outlines the full computation flow of the UniCast framework.
Visual and textual inputs are first tokenized and summarized via mean pooling, and a lightweight modality-aware context distiller extracts instance-specific contextual representations for each modality. This context is used to generate conditional soft prompts for the frozen vision and text encoders, producing contextualized visual and textual embeddings without updating encoder parameters.
The resulting multimodal embeddings are projected into the time-series representation space and used to generate an instance-conditioned time-series prompt. Through modality routing, auxiliary modality embeddings are selectively injected into the time-series patch representations via cross-attention, enabling input-dependent credit assignment. The combined sequence is then processed by a frozen TSFM with iterative routing, and forecasting is performed using the final TSFM output and a task-specific prediction head.

\begin{table*}[t]
    \caption[Vision Representation Strategies]{Comparison of different Vision embedding extraction strategies for UniCast. We compare using the First Token versus Average Pooling from CLIP and BLIP, both with and without the Qwen text model. For each dataset and model configuration, the best-performing strategy is highlighted in bold.}
    \label{tab:visionpooling}
{\fontsize{9}{11}\selectfont
    \begin{tabular}{| cc | cccc | cccc |}
    \hline
    \multirow{2}{*}{Vision} & \multirow{2}{*}{Text} & \multicolumn{4}{c|}{First Token} & \multicolumn{4}{c|}{Average Pooling}\\
    \cline{3-10}
    &  & NN5 & Aus & Tour & Avg & NN5 & Aus & Tour & Avg\\
    \hline
    CLIP & $\times$ & \textbf{0.4658} & \textbf{0.3450} & \textbf{1.3535} & \textbf{0.7214} & 0.4735 & 0.3760 & 1.3808 & 0.7434\\
    BLIP & $\times$ & \textbf{0.4625} & \textbf{0.3517} & \textbf{1.3559} & \textbf{0.7234} & 0.4717 & 0.3787 & 1.3750 & 0.7418\\
    CLIP & Qwen & \textbf{0.4635} & 0.3431 & \textbf{1.3573} & \textbf{0.7213} & 0.4671 & \textbf{0.3334} & 1.3720 & 0.7242\\
    BLIP & Qwen & \textbf{0.4601} & 0.3320 & \textbf{1.3499} & \textbf{0.7140} & 0.4641 & \textbf{0.3171} & 1.3764 & 0.7192\\
    \hline
    \end{tabular}
}
\end{table*}

\begin{table*}[t]
    \caption[Text Representation Strategies]{Comparison of different Text embedding extraction strategies for UniCast. We compare using the Last Token versus Average Pooling from Qwen and LLaMA, both with and without the CLIP vision model. For each dataset and model configuration, the best-performing strategy is highlighted in bold.}
    \label{tab:textpooling}
{\fontsize{9}{11}\selectfont
    \begin{tabular}{| cc | cccc | cccc |}
    \hline
    \multirow{2}{*}{Vision} & \multirow{2}{*}{Text} & \multicolumn{4}{c|}{Last Token} & \multicolumn{4}{c|}{Average Pooling}\\
    \cline{3-10}
    &  & NN5 & Aus & Tour & Avg & NN5 & Aus & Tour & Avg\\
    \hline
    $\times$ & Qwen & \textbf{0.6556} & 0.5786 & 1.4074 & 0.8805 & 0.6680 & \textbf{0.3753} & \textbf{1.4060} & \textbf{0.8164}\\
    $\times$ & LLaMA & 0.6662 & 0.3882 & \textbf{1.3567} & 0.8037 & \textbf{0.4706} & \textbf{0.3823} & 1.3680 & \textbf{0.7403}\\
    CLIP & Qwen & \textbf{0.4624} & \textbf{0.3304} & 1.3604 & \textbf{0.7177} & 0.4635 & 0.3431 & \textbf{1.3573} & 0.7213\\
    CLIP & LLaMA & 0.4597 & 0.3112 & \textbf{1.3499} & 0.7069 & \textbf{0.4558} & \textbf{0.3104} & 1.3534 & \textbf{0.7065}\\
    \hline
    \end{tabular}
}
\end{table*}

\begin{table}[t]
    \caption[Impact of Conditional Prompting and Modality Routing]{Impact of Conditional Prompting and Modality Routing. Conditional Prompting (CP) and Modality Routing (MR) are applied incrementally to analyze their individual and combined effects on performance along with zero-shot(ZS) and fine-tuning(FT) of time series backbone. \textbf{Bold} indicates the best performance and \underline{underlined} the second best for each dataset.}
    \label{tab:cpmrablation}
{\fontsize{9}{11}\selectfont
    \begin{tabular}{ccc| ccc |c}
    \hline
    \textbf{Chronos} & \textbf{CP} & \textbf{MR} & \textbf{NN5} & \textbf{Aus} & \textbf{Tour} & \textbf{Avg}\\
    \hline
    ZS &  &  & 0.7913 & 2.6701 & 2.4682 & 1.9765\\
    FT &  &  & \underline{0.4635} & 0.3606 & 1.3896 & 0.7379\\
    \multirow{4}{*}{UniCast} & $\times$ & $\times$ & 0.4677 & 0.3801 & 1.3950 & 0.7476\\
     & $\bigcirc$ & $\times$ & 0.4637 & 0.3600 & \underline{1.3553} & \underline{0.7263}\\
     & $\times$ & $\bigcirc$ & 0.4714 & \underline{0.3343} & 1.3910 & 0.7322\\
     & $\bigcirc$ & $\bigcirc$ & \textbf{0.4577} & \textbf{0.3059} & \textbf{1.3467} & \textbf{0.7034}\\
    \hline
    \end{tabular}
}
\end{table}

\section{Conditional Prompting and Modality Routing Ablation}
\label{app:cpmrablation}

This section provides the full results of the ablation studies on Conditional Prompting
(CP) and Modality Routing (MR). We report performance across multiple datasets for
different implementation choices, including static versus CP and heuristic versus MR. 

While the main text presents only a summary of these results using bar plots for clarity,
Table~\ref{tab:cpmrablation} shows the complete numerical results, allowing a detailed
comparison of each design choice and its effect on UniCast performance.

\begin{table}[t]
    \caption[Implementation of Conditional Prompting and Modality Routing]{Implementation of Conditional Prompting (CP) and Modality Routing (MR). CP compares Transformer (T) and MLP (M), while MR compares Shared (S) and Layerwise (L). \textbf{Bold} indicates the best performance and \underline{underlined} the second best for each dataset. The default configuration used in UniCast is Transformer for CP and Shared for MR.}
    \label{tab:cpmrimplementation}
{\fontsize{9}{11}\selectfont
    \begin{tabular}{cc| ccc |c}
    \hline
    \textbf{CP} & \textbf{MR} & \textbf{NN5} & \textbf{Aus-Elec} & \textbf{Tourism} & \textbf{Avg}\\
    \hline
    \multicolumn{2}{c|}{ZS} & 0.7913 & 2.6701 & 2.4682 & 1.9765\\
    \multicolumn{2}{c|}{FT} & \underline{0.4635} & 0.3606 & 1.3896 & 0.7379\\
    \hline
    T & S & \underline{0.4635} & 0.3431 & \textbf{1.3573} & \underline{0.7213}\\
    M & S & 0.4659 & \underline{0.3339} & 1.3786 & 0.7261\\
    T & L & \textbf{0.4627} & \textbf{0.3099} & \underline{1.3726} & \textbf{0.7151}\\
    \hline
    \end{tabular}
}
\end{table}

\section{Implementation Details of Conditional Prompting and Modality Routing}
\label{app:cpmrimplementation}

\textbf{Conditional Prompting.}
To investigate whether explicitly modeling interactions among multimodal inputs is necessary when constructing CP, we compare two implementations of CP, an MLP-based generator and a Transformer-based generator.
As shown in the Table~\ref{tab:cpmrimplementation}, the Transformer-based implementation consistently outperforms the MLP-based variant, indicating that capturing cross-modal dependencies during prompt generation is crucial for effective conditional prompting.
This suggests that CP benefits from architectures that can model structured interactions across modalities, rather than relying on simple aggregation mechanisms.
Based on these results, we adopt the Transformer-based CP as the default design in UniCast.

\textbf{Modality Routing.}
We evaluate two MR strategies: a Layerwise routing scheme, where cross-modal interactions are modeled at each transformer layer, and a Shared routing scheme, where a single routing module is shared across layers.
While the Layerwise MR achieves better performance, it introduces additional parameters and computational overhead.
In contrast, the Shared MR provides competitive performance with significantly improved parameter efficiency.
Considering UniCast’s goal of a parameter-efficient multimodal framework, we adopt Shared MR as the default configuration to balance performance and parameter efficiency.

\begin{figure*}[t]
\centering

\begin{subfigure}{\textwidth}
\centering
\includegraphics[width=0.24\linewidth]{figure/vision_heatmap_blip_nn5_1.png}
\includegraphics[width=0.24\linewidth]{figure/vision_heatmap_blip_nn5_2.png}
\includegraphics[width=0.24\linewidth]{figure/vision_heatmap_blip_nn5_3.png}
\includegraphics[width=0.24\linewidth]{figure/vision_heatmap_blip_nn5_4.png}
\\
\includegraphics[width=0.24\linewidth]{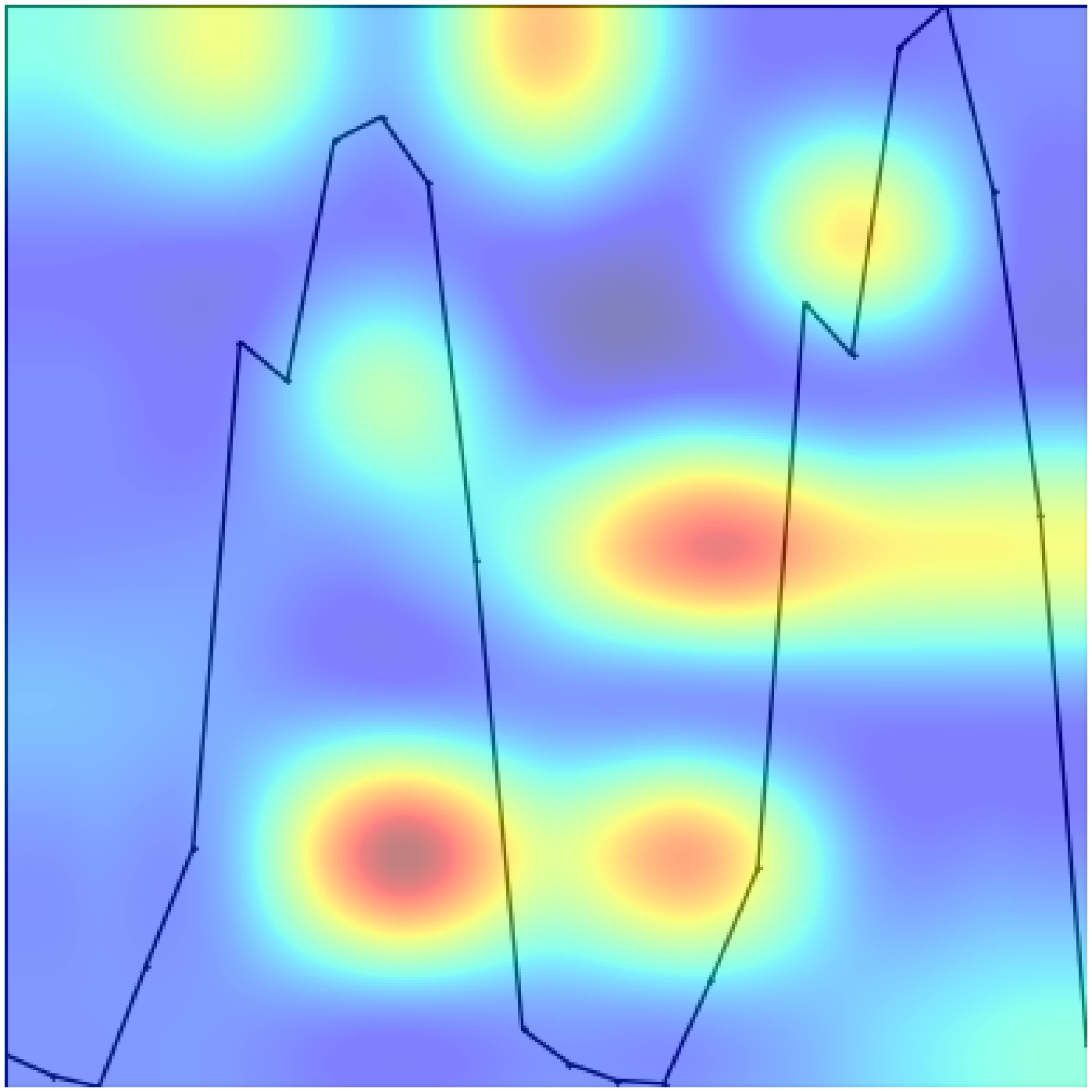}
\includegraphics[width=0.24\linewidth]{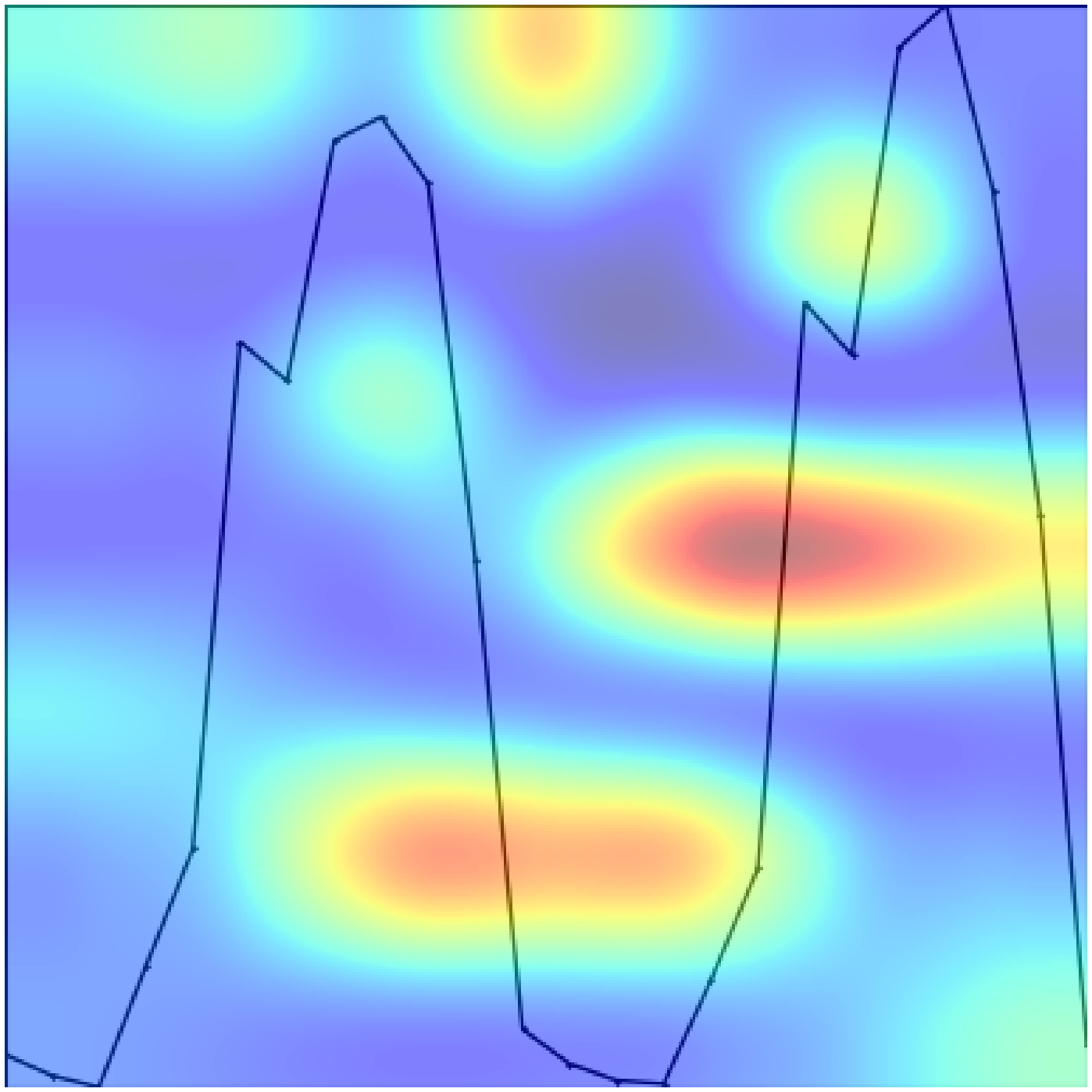}
\includegraphics[width=0.24\linewidth]{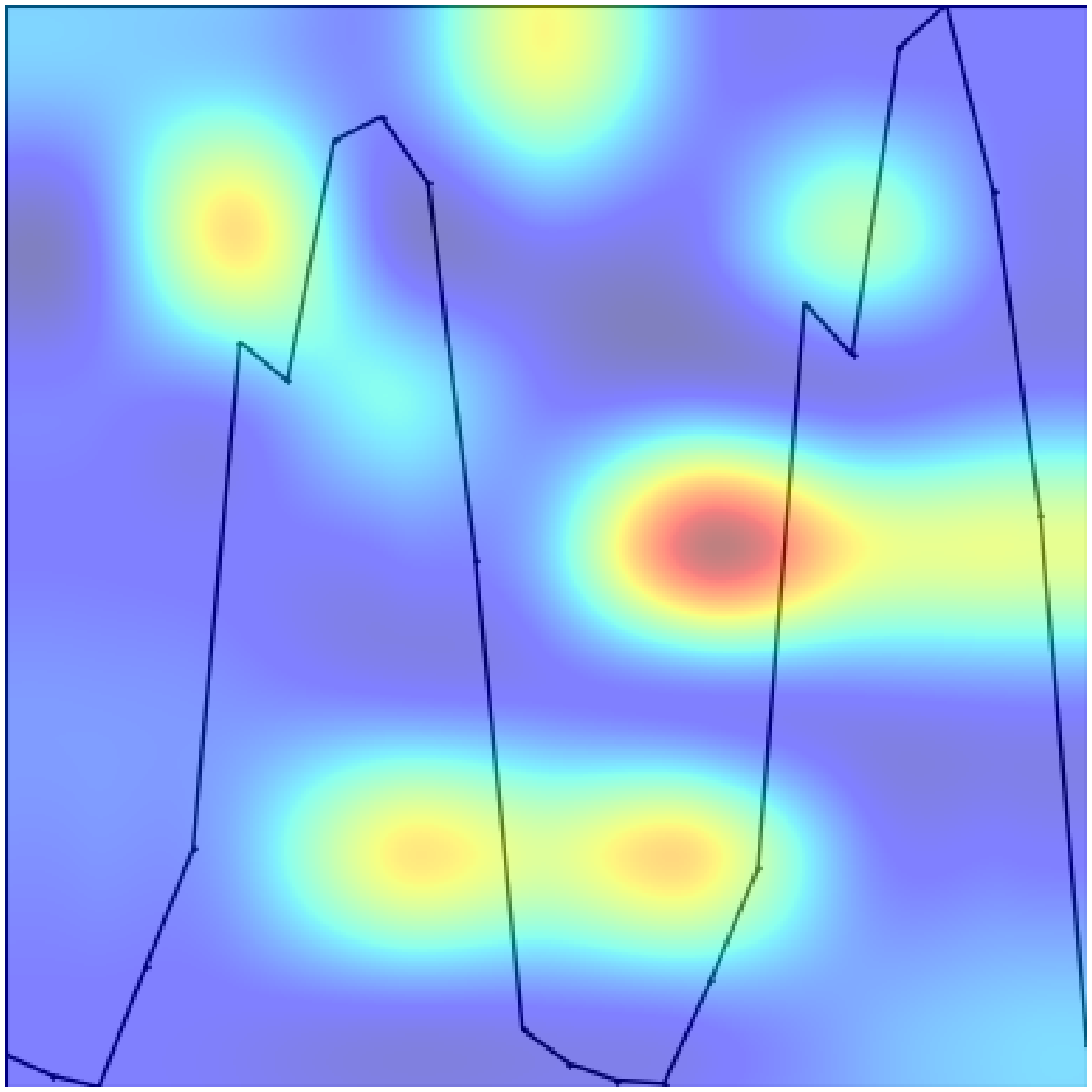}
\includegraphics[width=0.24\linewidth]{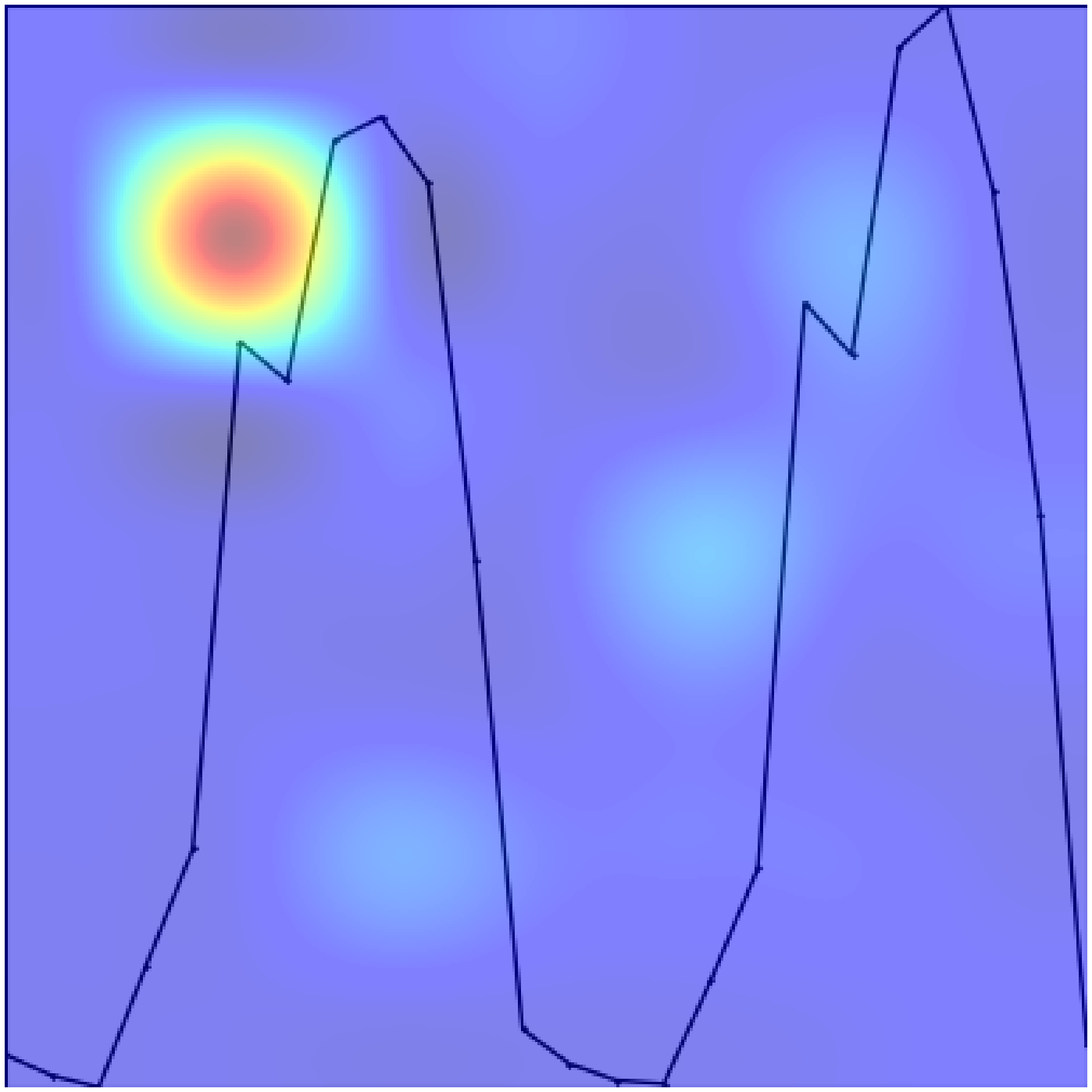}
\caption{BLIP on NN5 Daily and Tourism Monthly}
\end{subfigure}

\vspace{4pt}

\begin{subfigure}{\textwidth}
\centering
\includegraphics[width=0.24\linewidth]{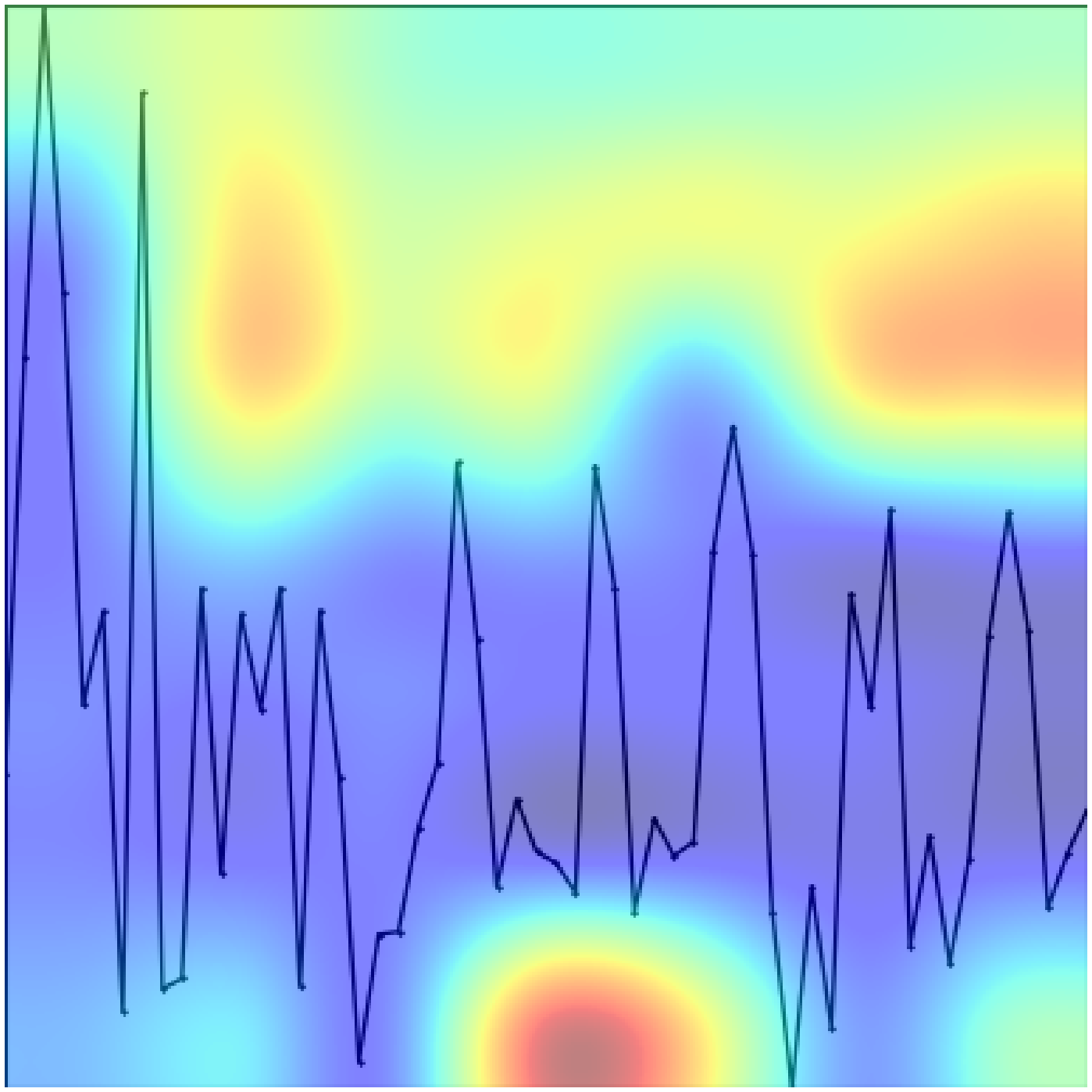}
\includegraphics[width=0.24\linewidth]{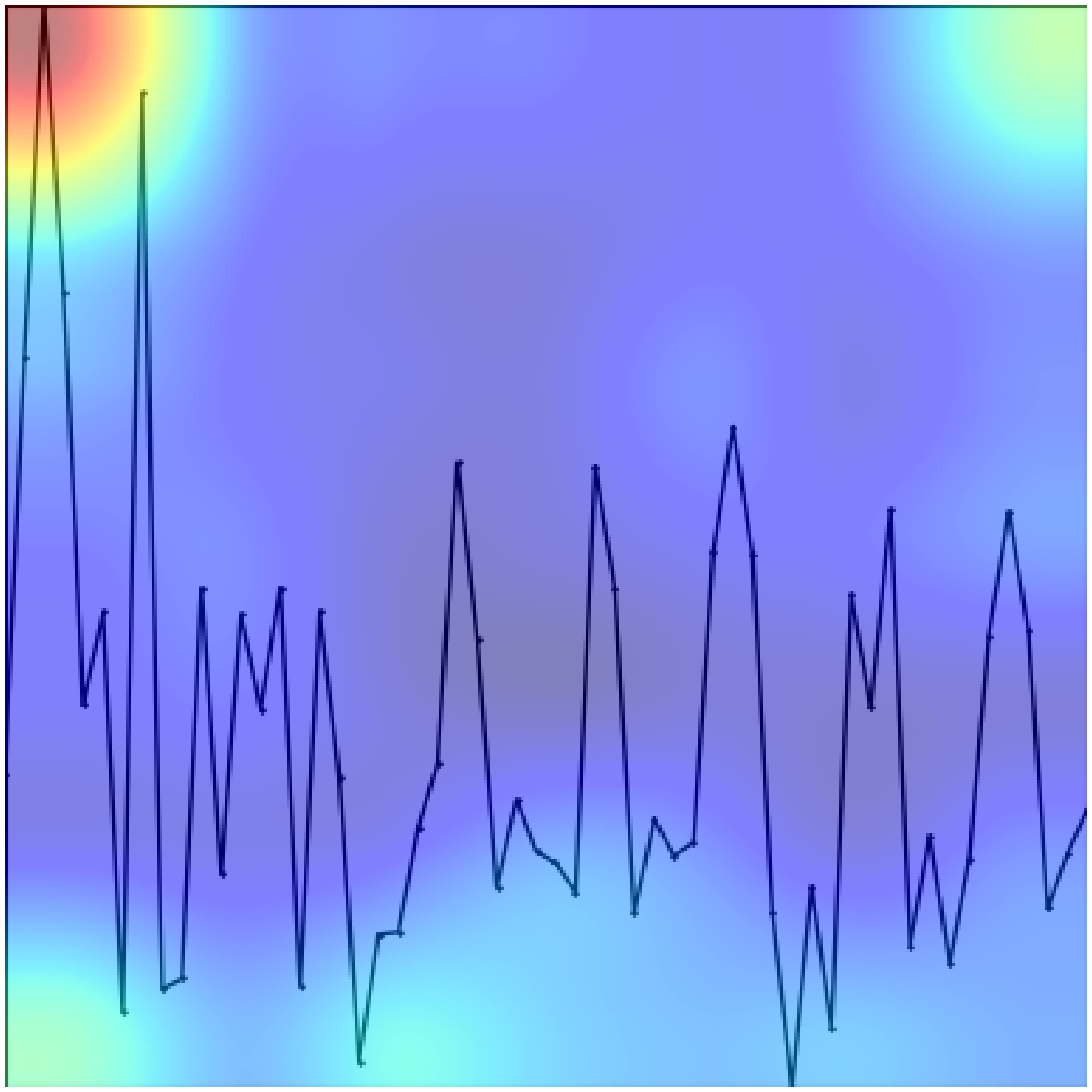}
\includegraphics[width=0.24\linewidth]{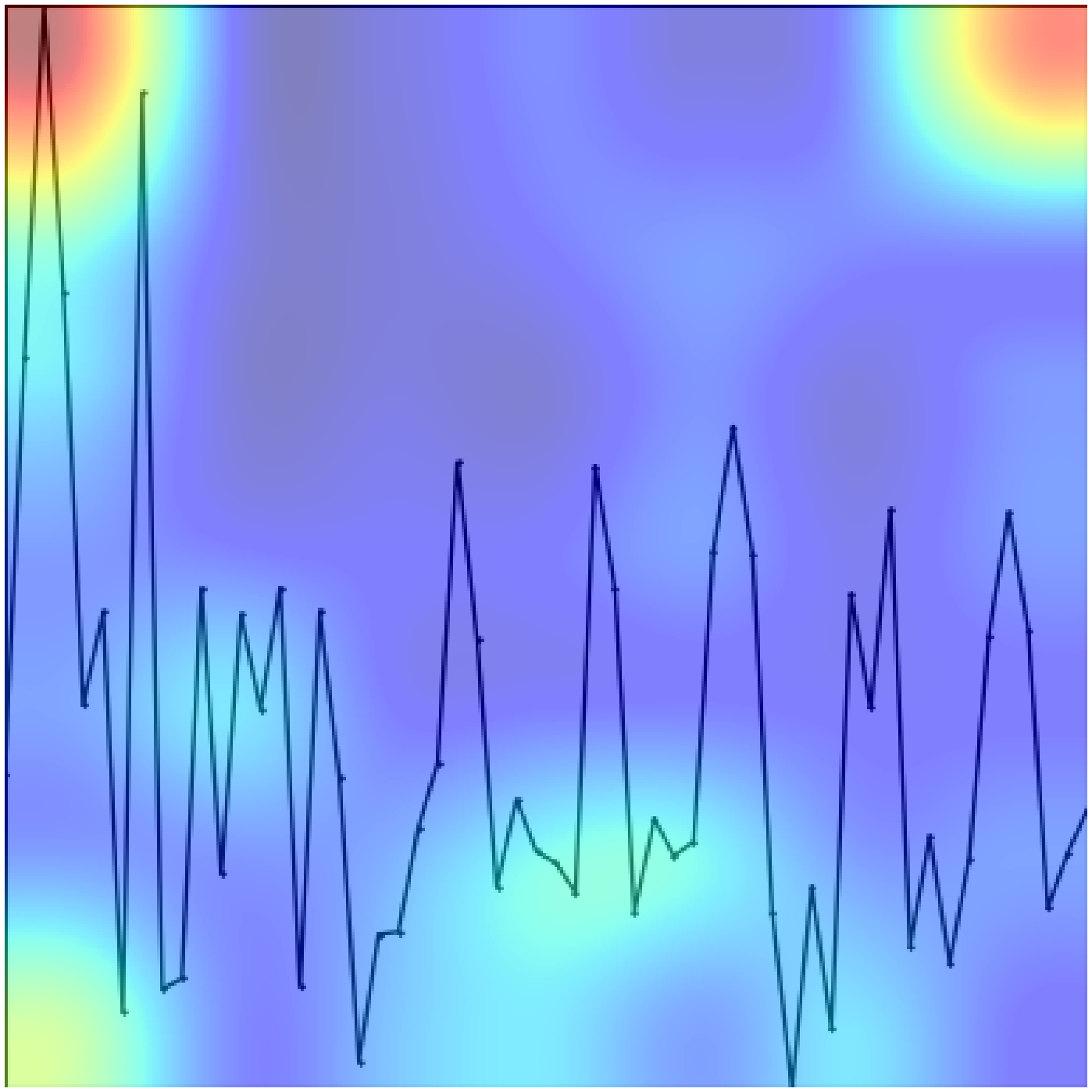}
\includegraphics[width=0.24\linewidth]{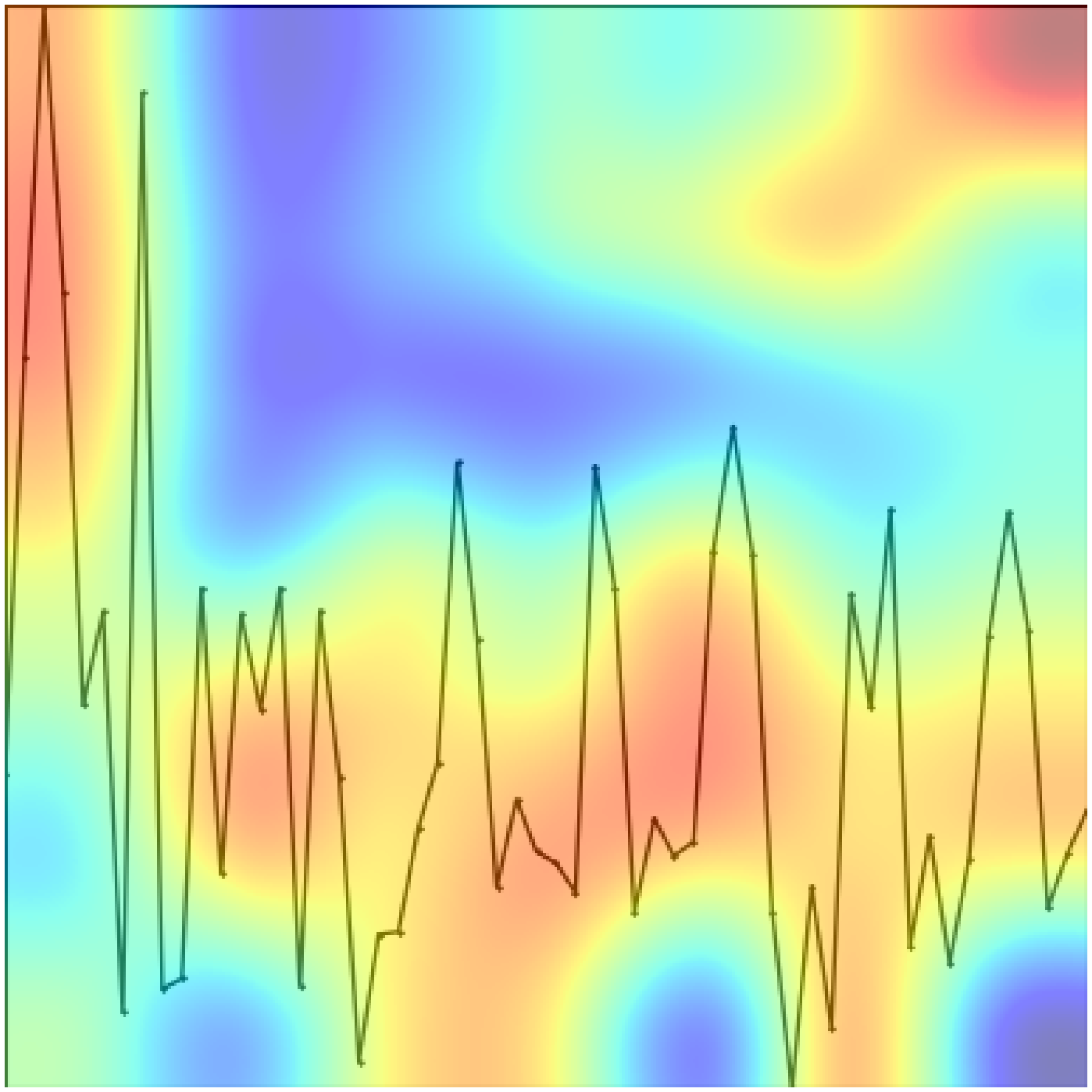}
\\
\includegraphics[width=0.24\linewidth]{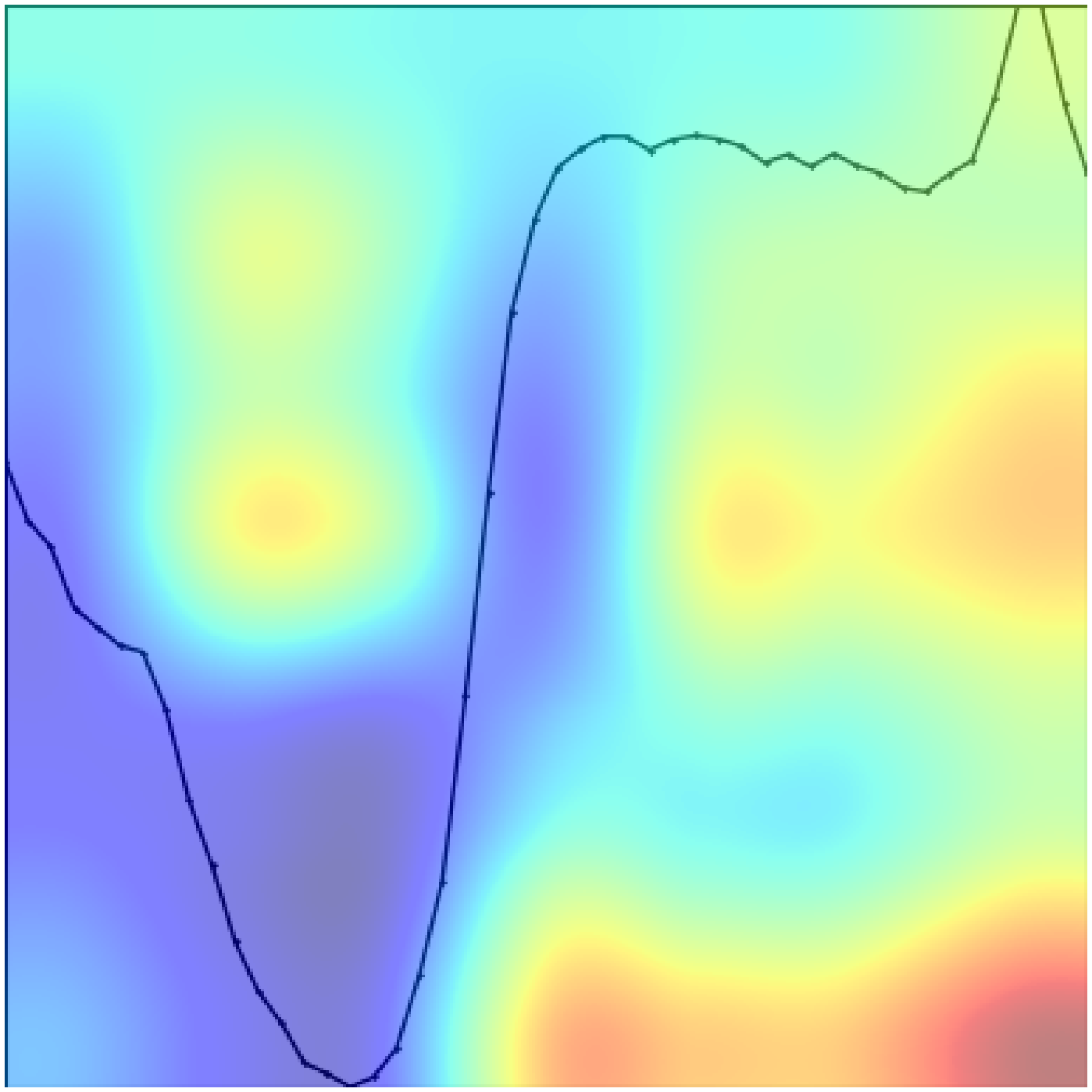}
\includegraphics[width=0.24\linewidth]{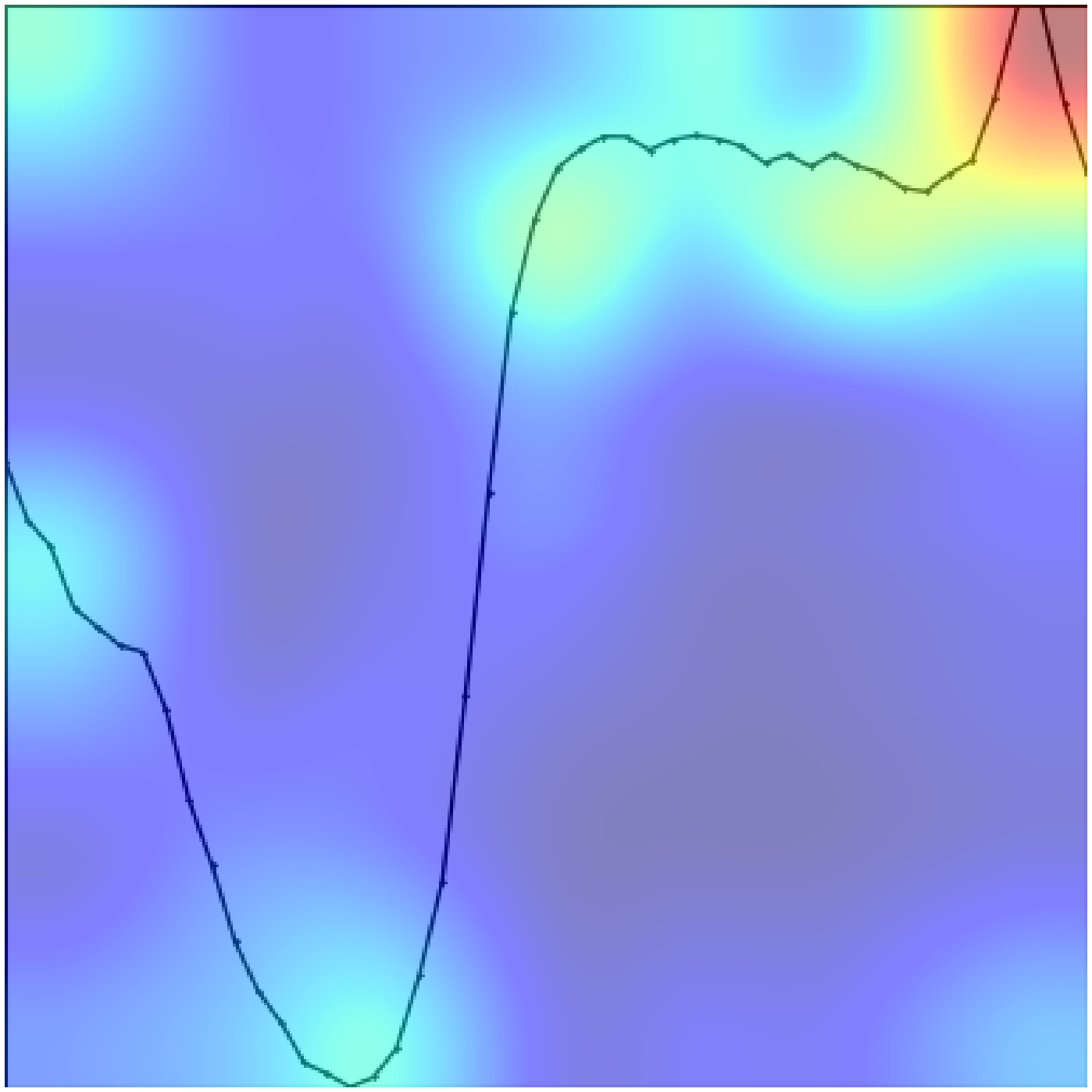}
\includegraphics[width=0.24\linewidth]{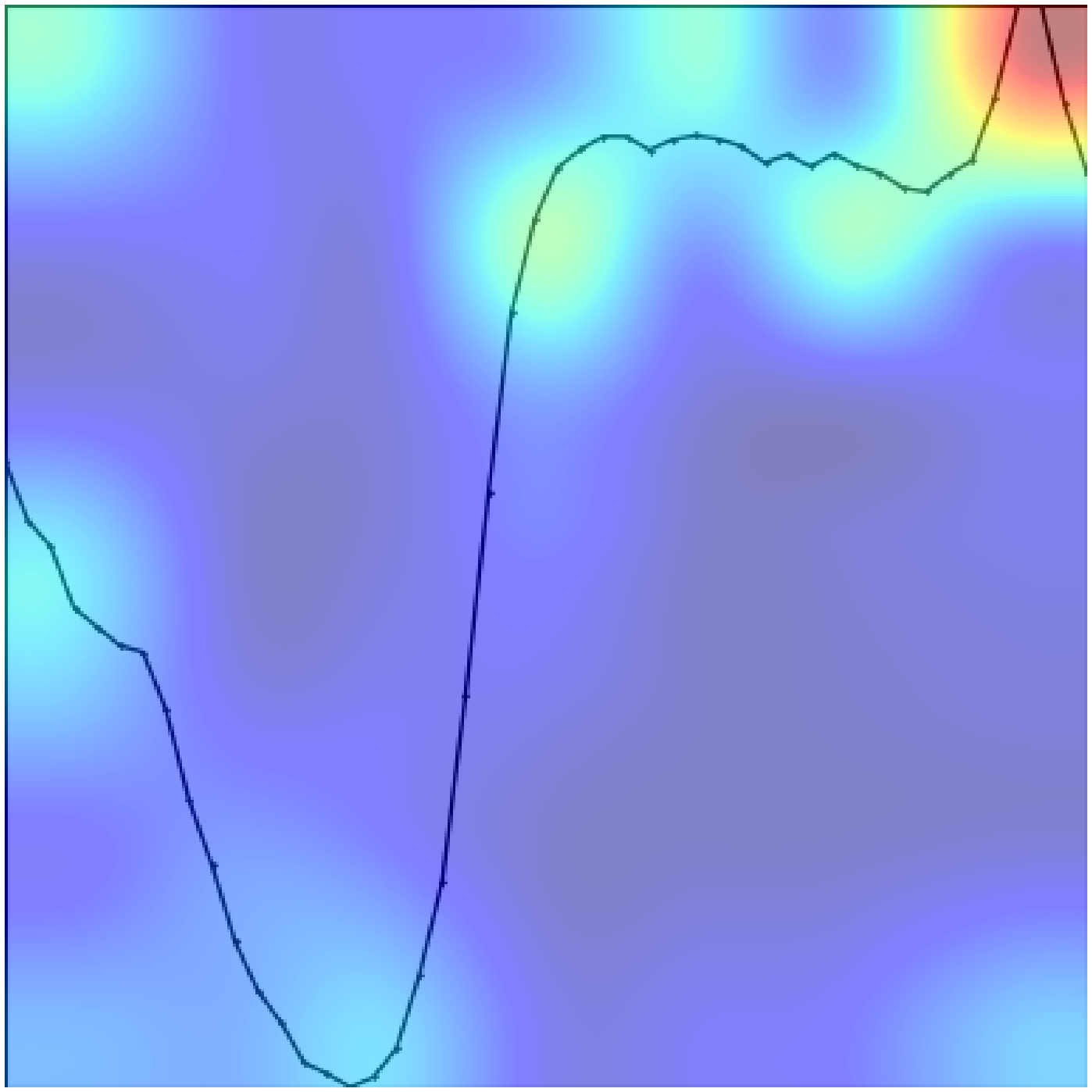}
\includegraphics[width=0.24\linewidth]{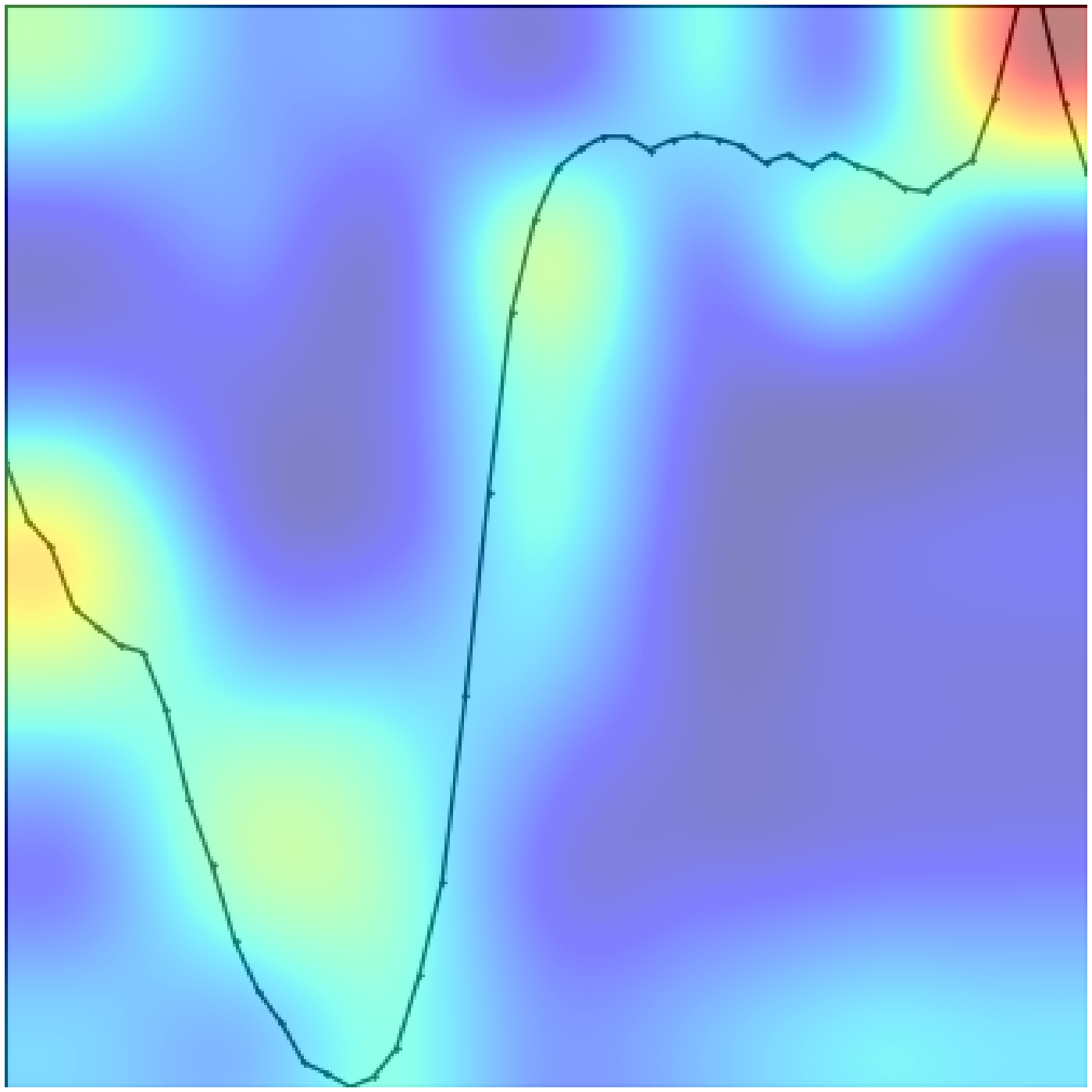}
\caption{CLIP on NN5 Daily and Australian Electricity}
\end{subfigure}

\caption[Vision Attention Heatmaps]{Vision Attention Heatmaps. For each subfigure, the feature map is split into four panels showing the attention at different layers from early to later layers.}
\label{fig:visionheatmap}
\end{figure*}

\begin{figure*}[t]
\centering

\begin{subfigure}{\textwidth}
\centering
\includegraphics[width=0.24\linewidth]{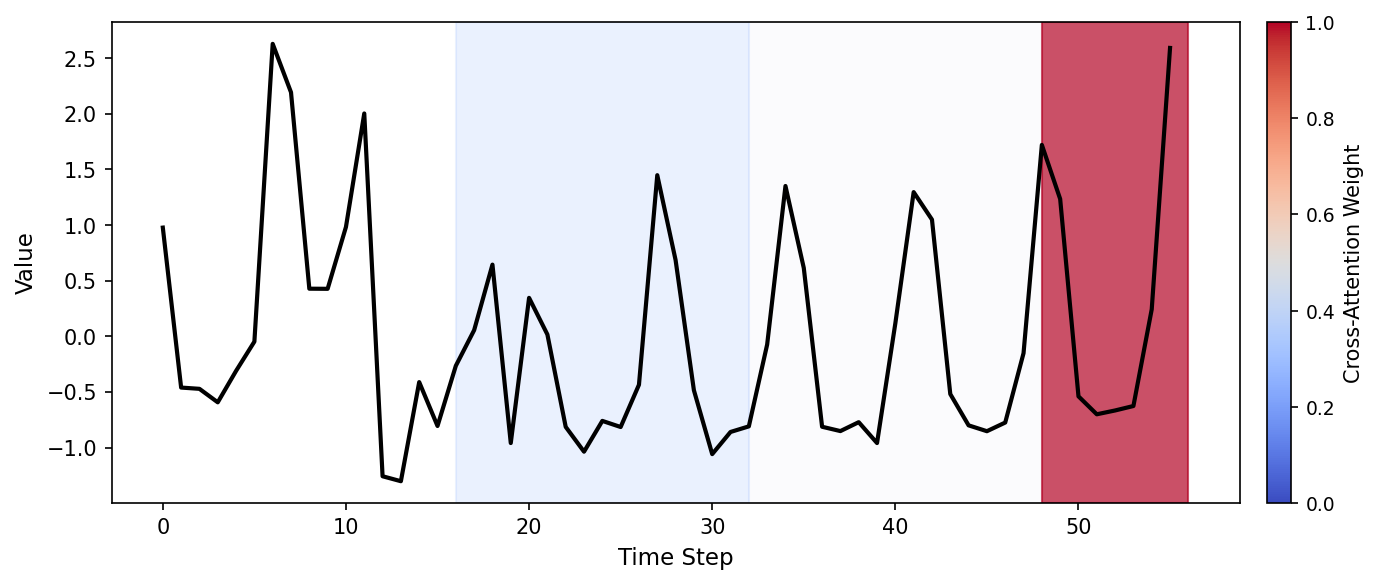}
\includegraphics[width=0.24\linewidth]{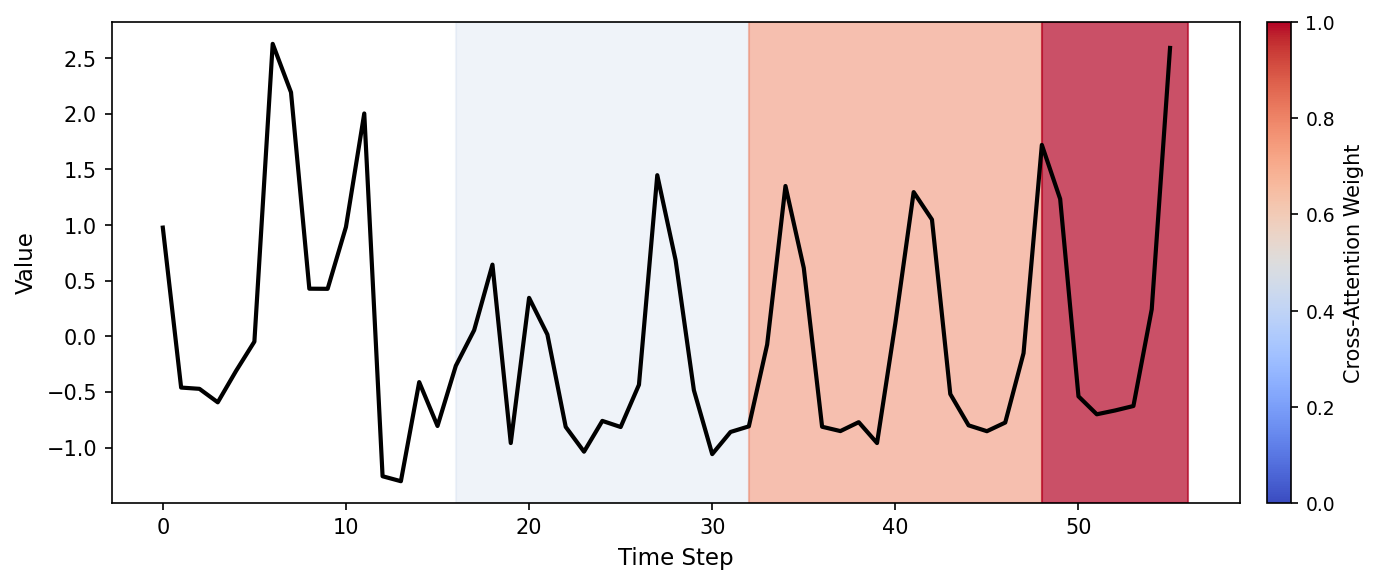}
\includegraphics[width=0.24\linewidth]{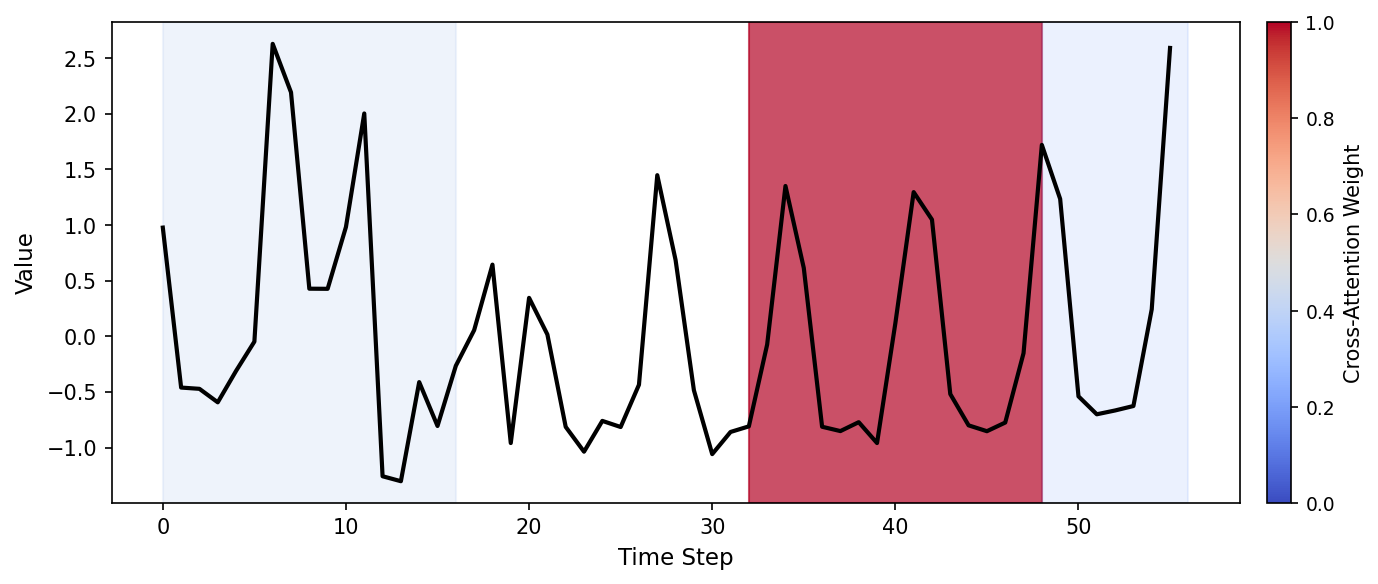}
\includegraphics[width=0.24\linewidth]{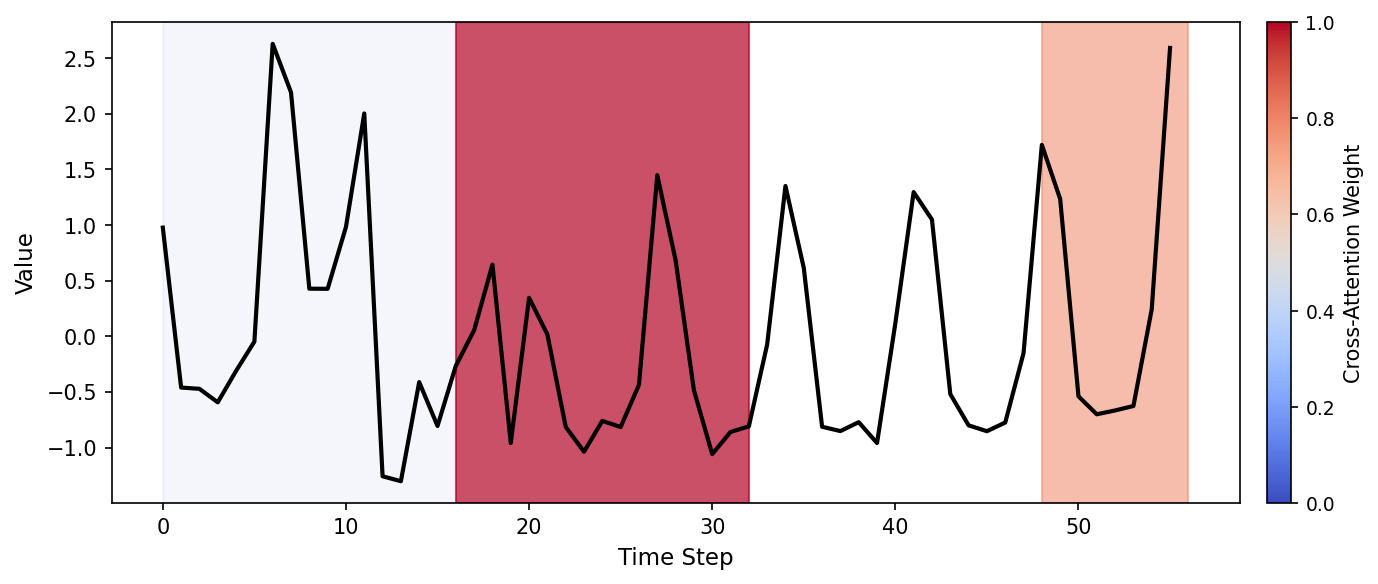}
\caption{}
\end{subfigure}

\vspace{4pt}

\begin{subfigure}{\textwidth}
\centering
\includegraphics[width=0.24\linewidth]{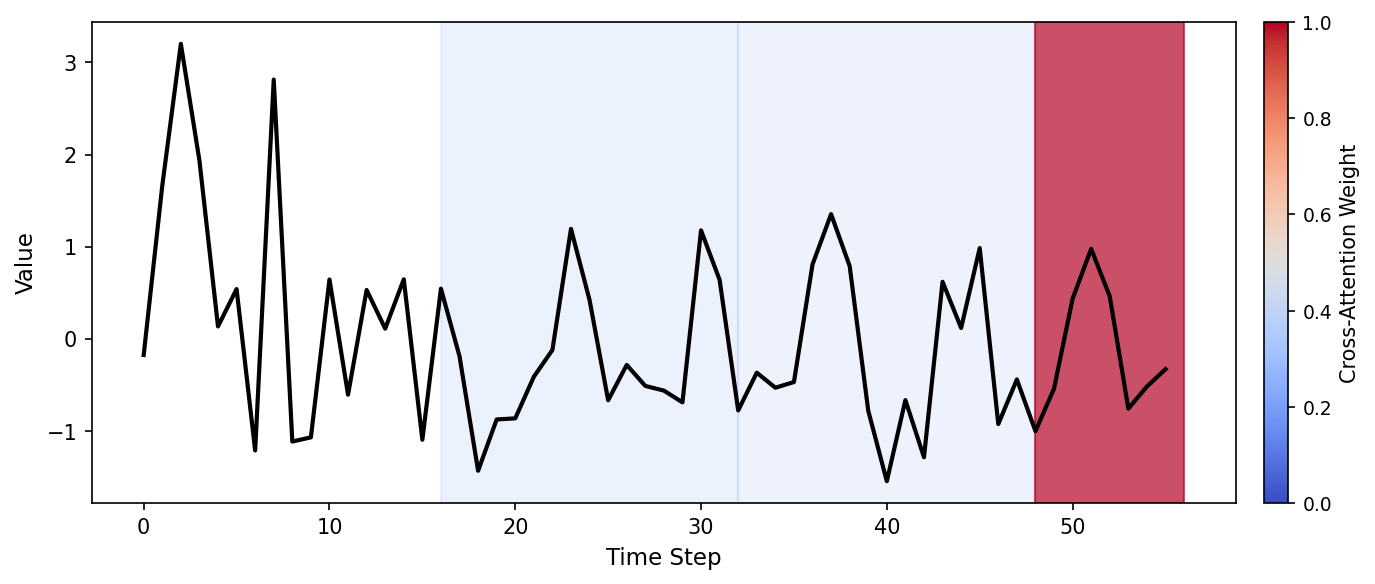}
\includegraphics[width=0.24\linewidth]{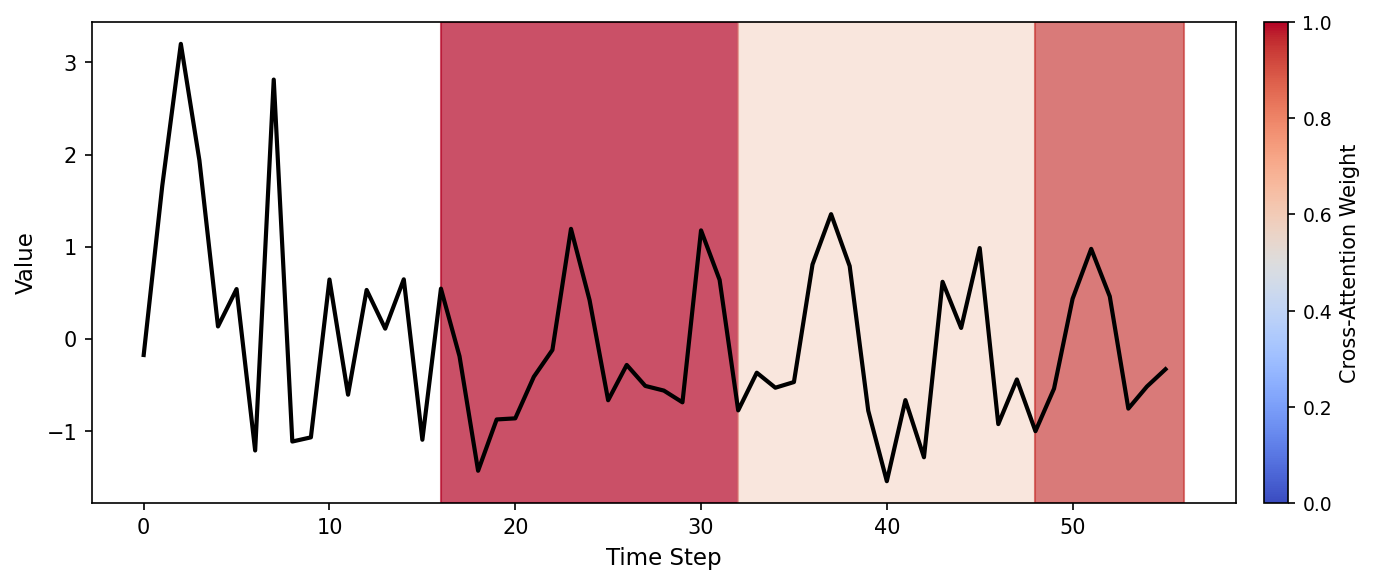}
\includegraphics[width=0.24\linewidth]{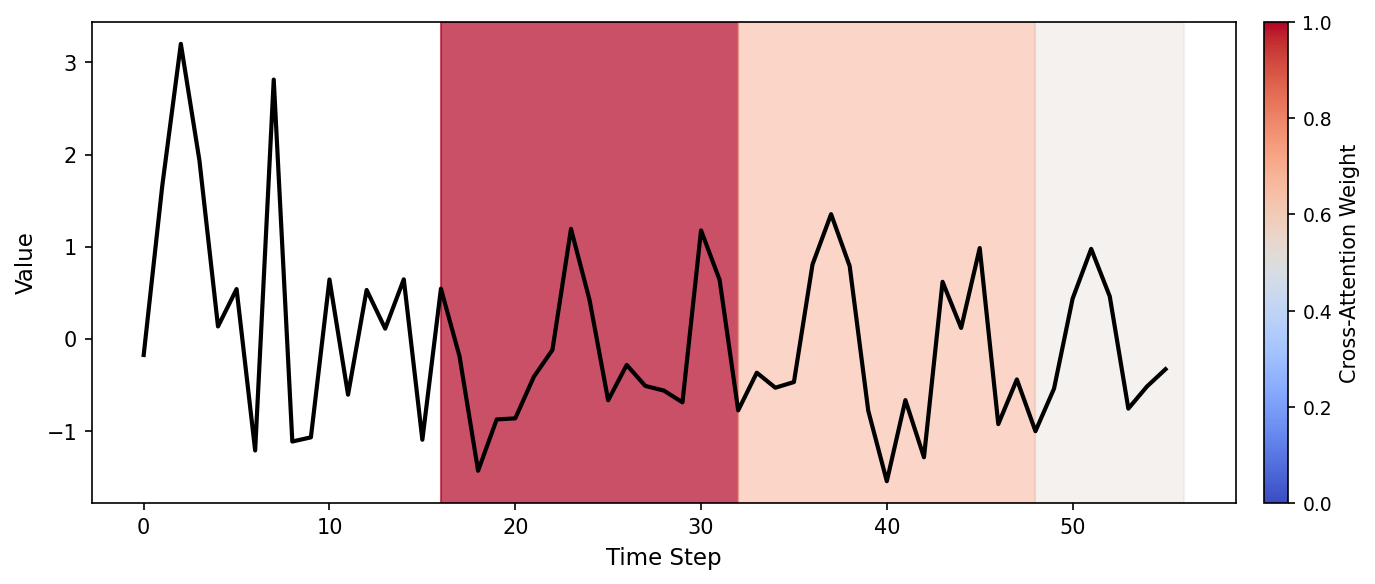}
\includegraphics[width=0.24\linewidth]{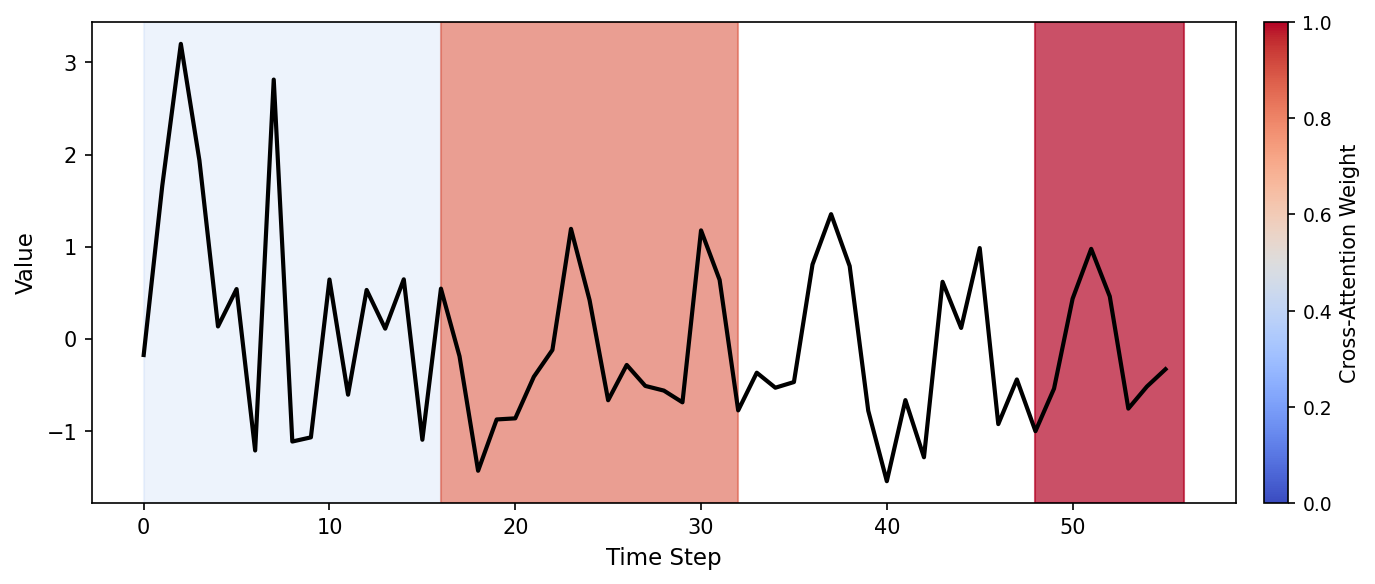}
\caption{}
\end{subfigure}

\vspace{4pt}

\begin{subfigure}{\textwidth}
\centering
\includegraphics[width=0.24\linewidth]{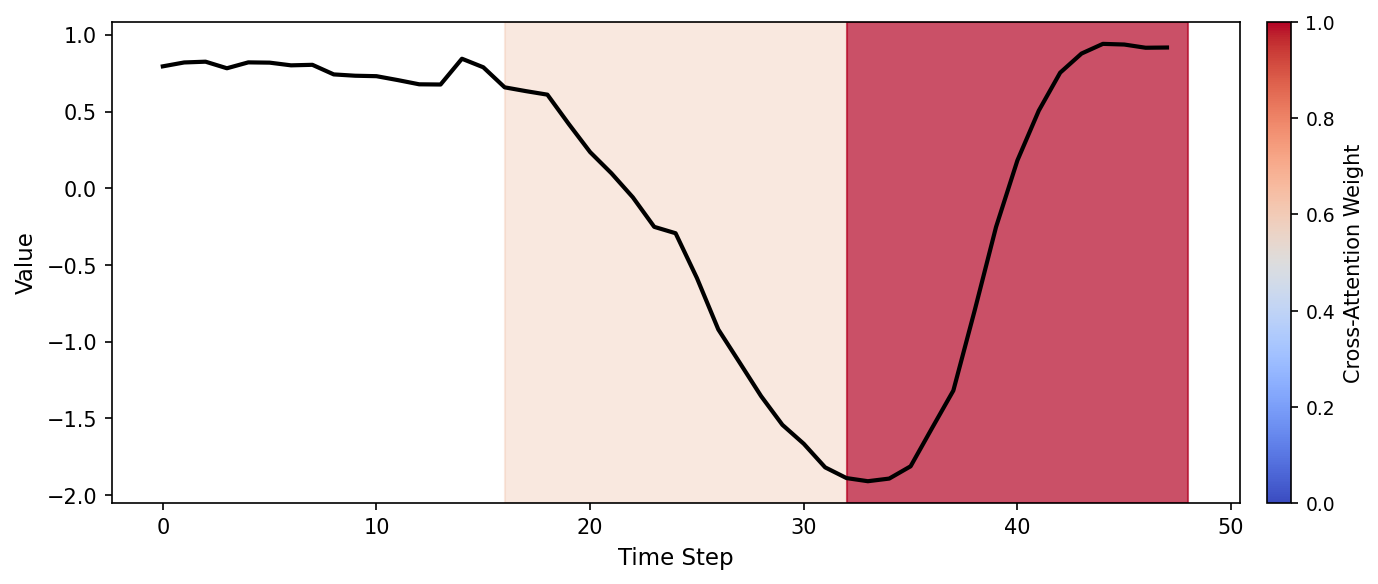}
\includegraphics[width=0.24\linewidth]{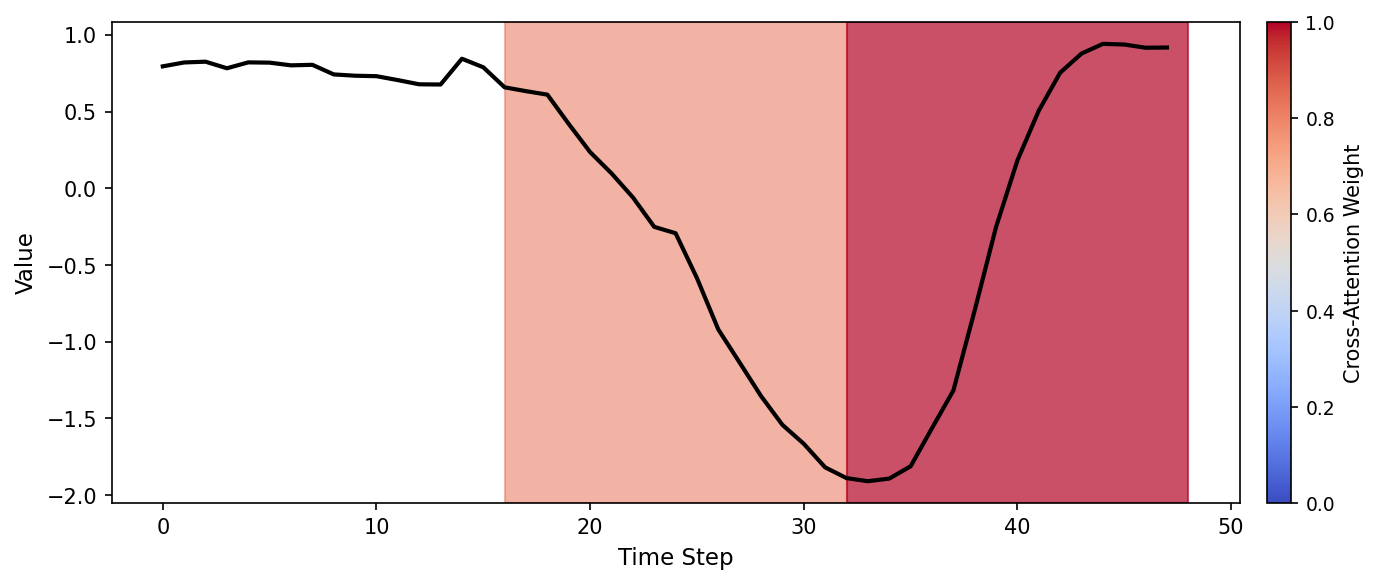}
\includegraphics[width=0.24\linewidth]{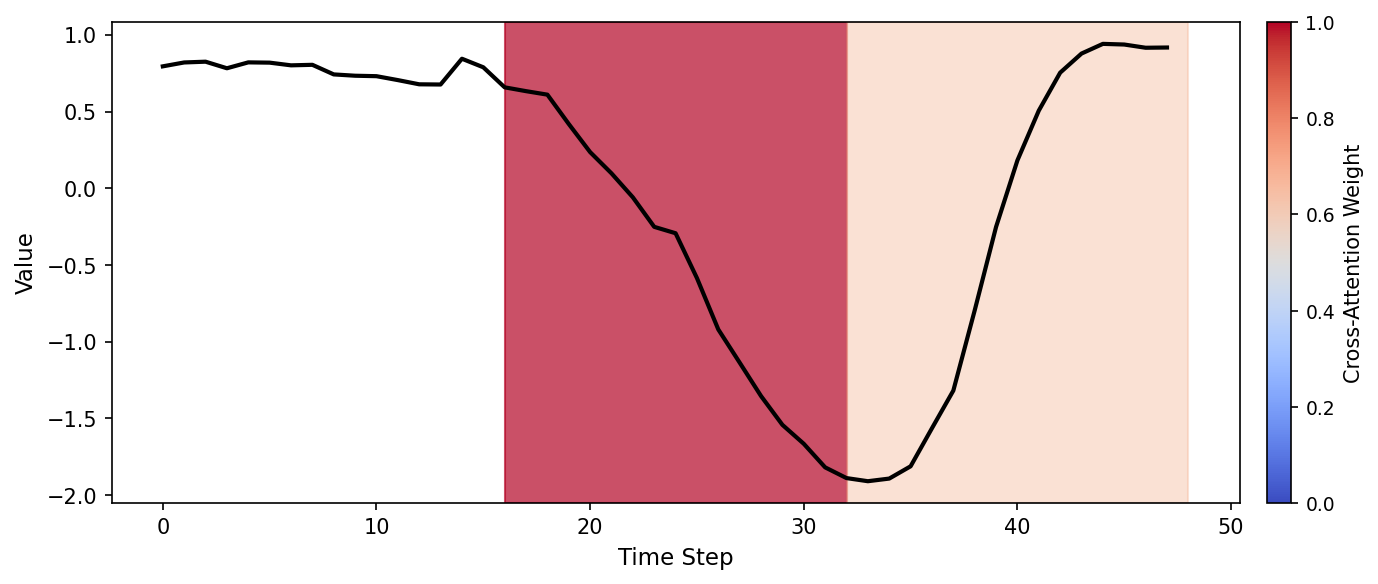}
\includegraphics[width=0.24\linewidth]{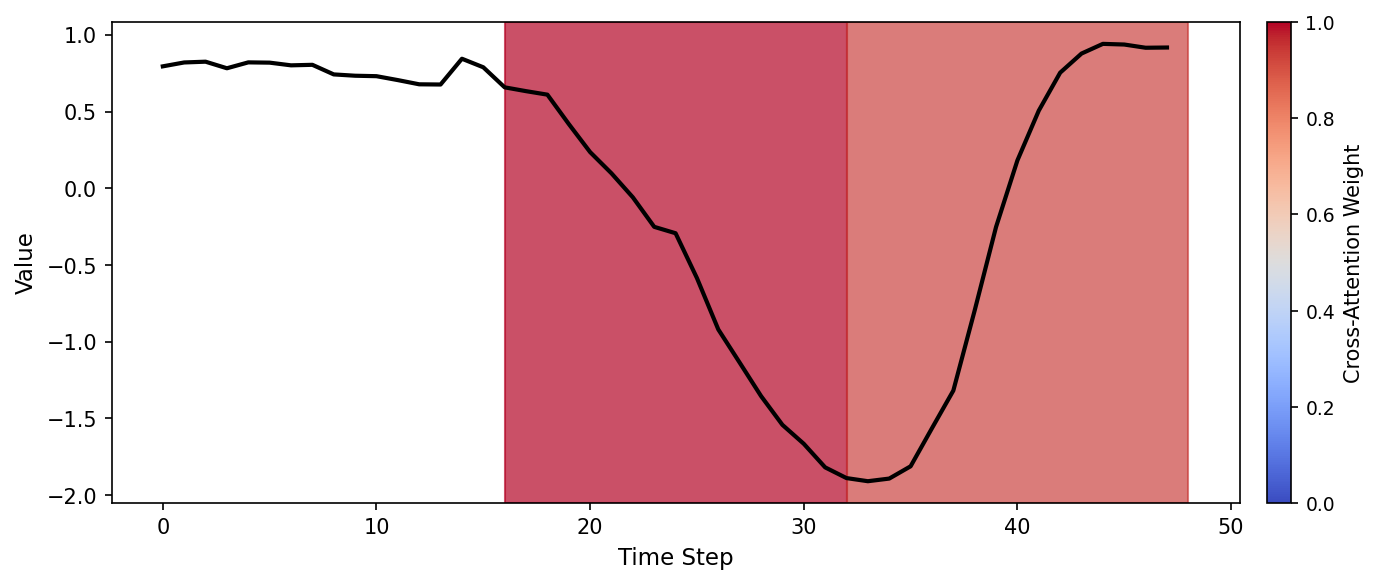}
\caption{}
\end{subfigure}

\caption[Time-series attention heatmaps]{Time-series attention heatmaps. These visualizations show the attention of Chronos to input time-series patches. Subfigures (a) and (b) correspond to the NN5 Daily dataset, while subfigure (c) corresponds to the Australian Electricity dataset. Each subfigure is split into four panels, showing attention at different layers from early to later layers, illustrating how the model progressively refines its focus across the network.}
\label{fig:timeseriesheatmap}
\vspace{-0.3cm}
\end{figure*}

\begin{figure}[t]
\centering

\begin{subfigure}{\columnwidth}
\centering
\raisebox{-0.5\height}{\includegraphics[width=0.55\linewidth]{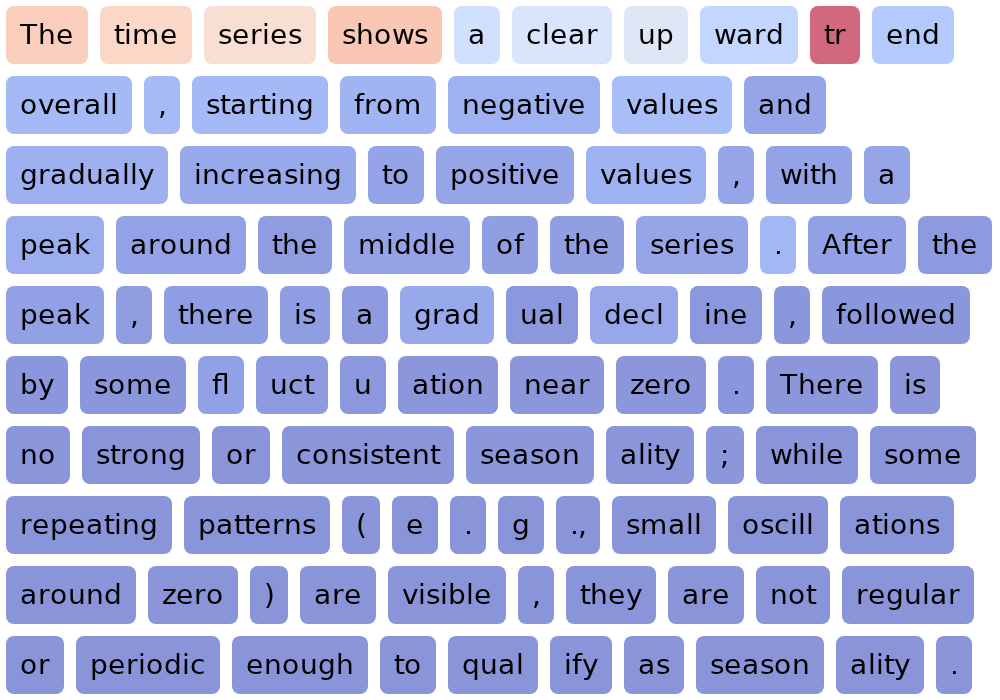}}
\hfill
\raisebox{-0.5\height}{\includegraphics[width=0.42\linewidth]{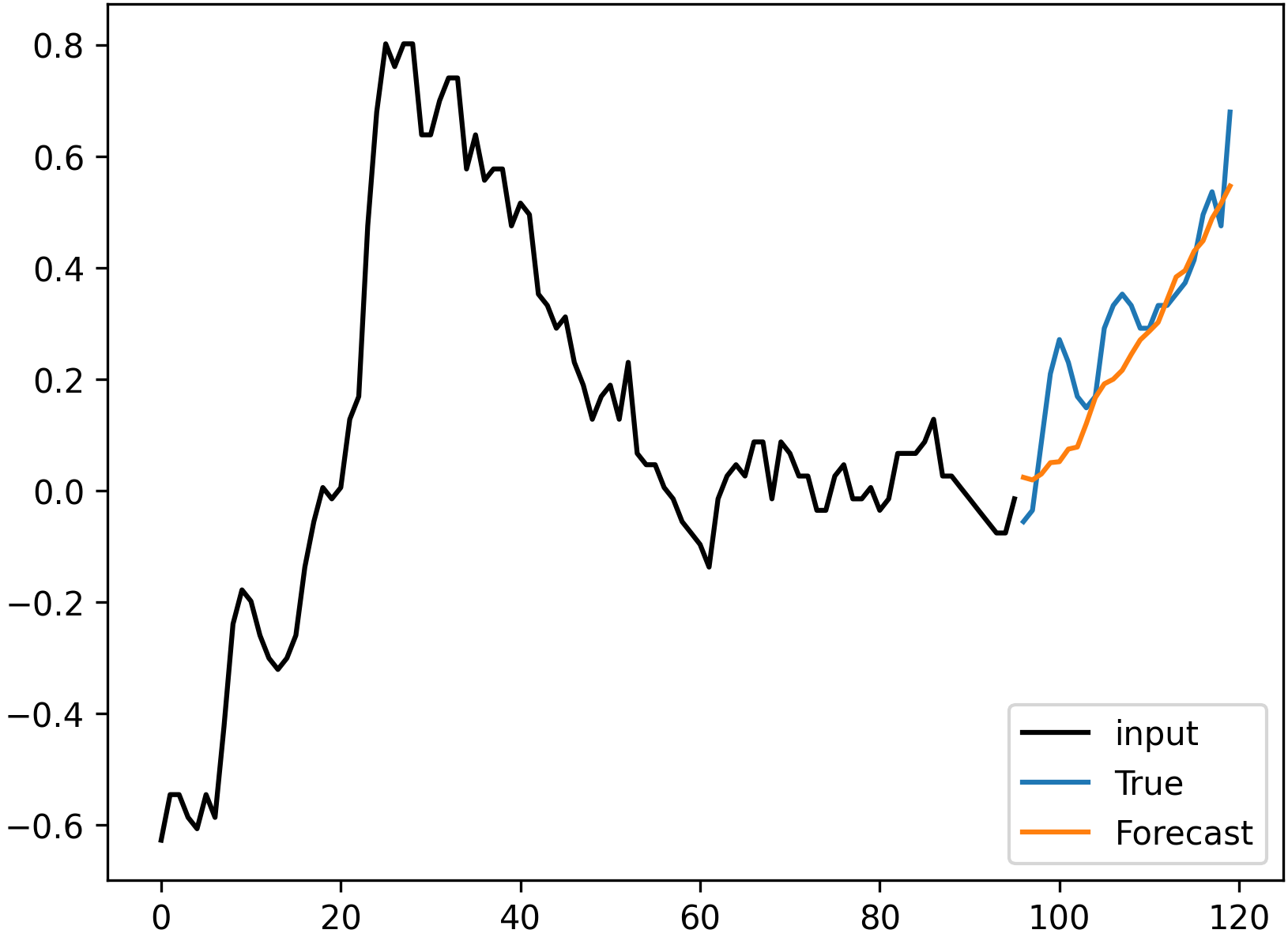}}
\caption{ETTm1}
\end{subfigure}

\vspace{4pt}

\begin{subfigure}{\columnwidth}
\centering
\raisebox{-0.5\height}{\includegraphics[width=0.55\linewidth]{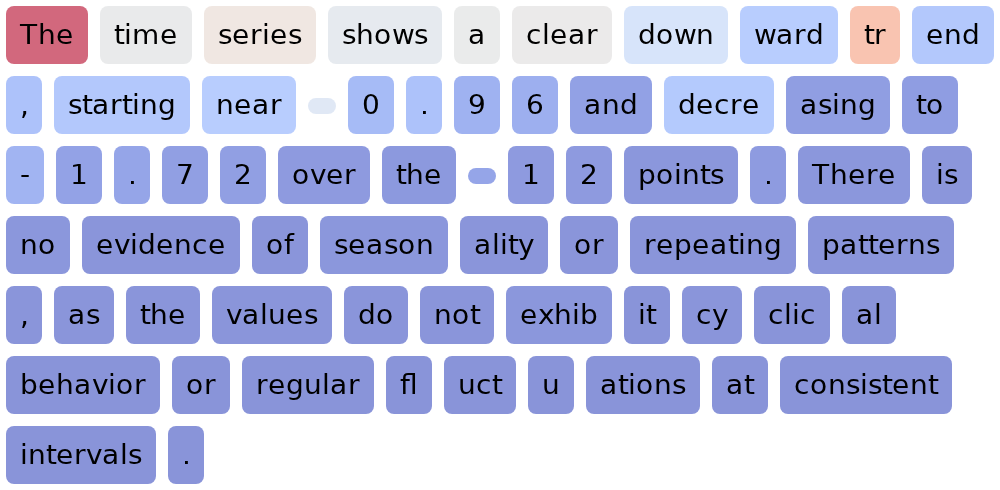}}
\hfill
\raisebox{-0.5\height}{\includegraphics[width=0.42\linewidth]{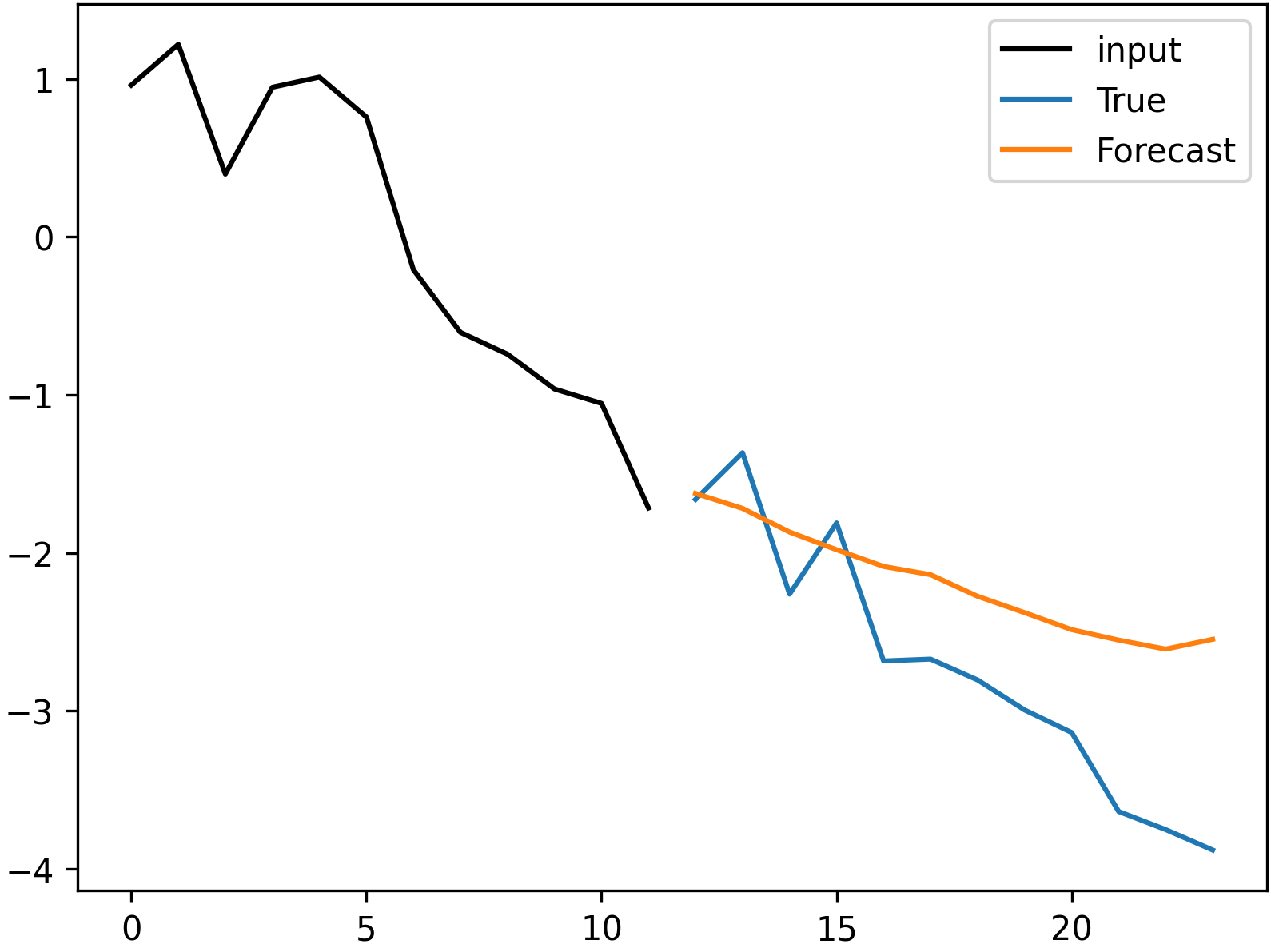}}
\caption{CIF 2016}
\end{subfigure}

\caption[Text attention heatmaps]{Text attention heatmaps. For each dataset, the left panel displays the attention heatmap across all layers of the text model for the corresponding textual description, while the right panel shows the forecast results. These visualizations illustrate how the text model attends to different parts of the input and how it contributes to the prediction.}
\label{fig:textheatmap}
\vspace{-0.3cm}
\end{figure}

\section{Effective Multimodal Representation Strategies}
\label{app:representationstrategy}
To analyze the effect of embedding extraction strategies on UniCast performance, we compare different pooling methods for Vision and Text modalities. For Vision backbones (CLIP, BLIP), we compare using the First Token versus Average Pooling, while for Text backbones (Qwen, LLaMA), we compare using the Last Token versus Average Pooling.

\textbf{Vision.}
For vision encoders, using the First Token consistently outperforms Average Pooling across both CLIP and BLIP backbones as shown in Table~\ref{tab:visionpooling}.
This result aligns with the design of vision transformer–based encoders, where the first token serves as a global representation that aggregates information from all image patches through self-attention.
During pretraining, both CLIP and BLIP optimize this token to capture high-level semantic information for cross-modal alignment, making it particularly suitable as a compact representation for multimodal conditioning.
In contrast, Average Pooling treats all patch tokens equally, which may dilute salient visual features that are most relevant for forecasting.
These findings suggest that leveraging the pretrained global representation token is more effective than uniform spatial aggregation for the Vision modality.
Based on this observation, we adopt the First Token as the default embedding extraction strategy for Vision in UniCast.

\textbf{Text.}
Table~\ref{tab:textpooling} reports that Average Pooling consistently outperforms using the Last Token representation when decoder-only language models (Qwen and LLaMA) are used as text backbones .
This behavior can be attributed to the architectural characteristics of decoder-only transformers, where the final token representation is optimized for next-token prediction during pretraining.
As a result, the last token embedding may reflect generation-oriented context rather than a balanced summary of the entire textual input.
In contrast, Average Pooling aggregates information across all tokens, producing a more stable sequence-level semantic representation that better captures forecasting-relevant context.
This suggests that, when decoder-only language models are used for multimodal conditioning rather than text generation, global token aggregation is more suitable than relying on the final token representation.
Based on this observation, we adopt Average Pooling as the default embedding extraction strategy for the Text modality in UniCast.

\begin{table*}[t]
    \centering
    \caption[Parameter Efficiency]{Comparison of parameter efficiency across UniCast configurations, including total parameters, trainable parameters, their ratio, and relative parameter percentage with respect to the time-series backbone Chronos. Unless otherwise specified, CP and MR denote Transformer-based CP and Shared MR, respectively.}
    \label{tab:parameterefficiency}
{\fontsize{9}{11}\selectfont
    \begin{tabular}{c|cc|cccc}
        \hline
        & \textbf{Vision} & \textbf{Text} & \textbf{\# of Trainable Params} & \textbf{\# of Total Params} & \textbf{Trainable Ratio} & \textbf{Relative Params (\%)}\\
        \hline
        \hline
        \multirow{6}{*}{\rotatebox{90}{\textbf{Main}}}& \multicolumn{2}{c|}{\textbf{Chronos}} &205,292,928&205,292,928&100.00\%&100.00\%\\
        \cline{2-7}
        & \cellcolor{lightgray!30}\textbf{CP} &  \cellcolor{lightgray!30}\textbf{MR} & \cellcolor{lightgray!30} & \cellcolor{lightgray!30} & \cellcolor{lightgray!30} & \cellcolor{lightgray!30}\\
        \cline{2-7}
        & \textbf{CLIP} & \textbf{Qwen} & 115,356,672 & 1,951,819,904 & 5.91\% & 56.19\%\\
        & \textbf{BLIP} & \textbf{Qwen} & 115,356,672 & 1,950,454,400 & 5.91\% & 56.19\%\\
        & \textbf{CLIP} & \textbf{LLaMA} & 395,948,032 & 7,296,040,576 & 5.43\% & 192.87\%\\
        & \textbf{BLIP} & \textbf{LLaMA} & 395,948,032 & 7,294,675,072 & 5.43\% & 192.87\%\\
        \hline
        \hline
        \multirow{24}{*}{\rotatebox{90}{\textbf{Ablations}}} & \cellcolor{lightgray!30}\textbf{Static} & \cellcolor{lightgray!30}\textbf{Heuristic} & \cellcolor{lightgray!30} & \cellcolor{lightgray!30} & \cellcolor{lightgray!30} & \cellcolor{lightgray!30}\\
        \cline{2-7}
        & \textbf{CLIP} & \textbf{Qwen} & 2,072,064 & 1,838,535,296 & 0.11\% & 1.01\%\\
        & \textbf{BLIP} & \textbf{Qwen} & 2,072,064 & 1,837,169,792 & 0.11\% & 1.01\%\\
        & \textbf{CLIP} & \textbf{LLaMA} & 4,390,400 & 6,904,482,944 & 0.06\% & 2.14\%\\
        & \textbf{BLIP} & \textbf{LLaMA} & 4,390,400 & 6,903,117,440 & 0.06\% & 2.14\%\\
        \cline{2-7}
        & \cellcolor{lightgray!30}\textbf{CP} & \cellcolor{lightgray!30}\textbf{Heuristic} & \cellcolor{lightgray!30} & \cellcolor{lightgray!30} & \cellcolor{lightgray!30} & \cellcolor{lightgray!30}\\
        \cline{2-7}
        & \textbf{CLIP} & \textbf{Qwen} & 112,994,304 & 1,949,457,536 & 5.80\% & 55.04\%\\
        & \textbf{BLIP} & \textbf{Qwen} & 112,994,304 & 1,948,092,032 & 5.80\% & 55.04\%\\
        & \textbf{CLIP} & \textbf{LLaMA} & 393,585,664 & 7,293,678,208 & 5.40\% & 191.72\%\\
        & \textbf{BLIP} & \textbf{LLaMA} & 393,585,664 & 7,292,312,704 & 5.40\% & 191.72\%\\
        \cline{2-7}
        & \cellcolor{lightgray!30}\textbf{Static} & \cellcolor{lightgray!30}\textbf{MR} & \cellcolor{lightgray!30} & \cellcolor{lightgray!30} & \cellcolor{lightgray!30} & \cellcolor{lightgray!30}\\
        \cline{2-7}
        & \textbf{CLIP} & \textbf{Qwen} & 4,434,432 & 1,840,897,664 & 0.24\% & 2.16\%\\
        & \textbf{BLIP} & \textbf{Qwen} & 4,434,432 & 1,839,532,160 & 0.24\% & 2.16\%\\
        & \textbf{CLIP} & \textbf{LLaMA} & 6,752,768 & 6,906,845,312 & 0.10\% & 3.29\%\\
        & \textbf{BLIP} & \textbf{LLaMA} & 6,752,768 & 6,905,479,808 & 0.10\% & 3.29\%\\
        \cline{2-7}
        & \cellcolor{lightgray!30}\textbf{CP}(MLP) & \cellcolor{lightgray!30}\textbf{MR} & \cellcolor{lightgray!30} & \cellcolor{lightgray!30} & \cellcolor{lightgray!30} & \cellcolor{lightgray!30}\\
        \cline{2-7}
        & \textbf{CLIP} & \textbf{Qwen} & 40,717,056 & 1,877,180,288 & 2.17\% & 19.83\%\\
        & \textbf{BLIP} & \textbf{Qwen} & 40,717,056 & 1,875,814,784 & 2.17\% & 19.83\%\\
        & \textbf{CLIP} & \textbf{LLaMA} & 121,991,936 & 7,022,084,480 & 1.74\% & 59.42\%\\
        & \textbf{BLIP} & \textbf{LLaMA} & 121,991,936 & 7,020,718,976 & 1.74\% & 59.42\%\\
        \cline{2-7}
        \cline{2-7}
        & \cellcolor{lightgray!30}\textbf{CP} & \cellcolor{lightgray!30}\textbf{MR}(Layerwise) & \cellcolor{lightgray!30} & \cellcolor{lightgray!30} & \cellcolor{lightgray!30} & \cellcolor{lightgray!30}\\
        \cline{2-7}
        & \textbf{CLIP} & \textbf{Qwen} & 172,053,504 & 2,008,516,736 & 8.57\% & 83.81\%\\
        & \textbf{BLIP} & \textbf{Qwen} & 172,053,504 & 2,007,151,232 & 8.57\% & 83.81\%\\
        & \textbf{CLIP} & \textbf{LLaMA} & 452,644,864 & 7,352,737,408 & 6.16\% & 220.49\%\\
        & \textbf{BLIP} & \textbf{LLaMA} & 452,644,864 & 7,351,371,904 & 6.16\% & 220.49\%\\
        \bottomrule
    \end{tabular}
}
\end{table*}

\section{Heatmaps}
\label{app:heatmaps}
\textbf{Vision}

To qualitatively analyze how UniCast leverages visual information for time-series forecasting, we visualize layer-wise attention heatmaps from the vision modality.
Figure~\ref{fig:visionheatmap} shows the attention heatmaps of BLIP and CLIP, which are pretrained vision models for UniCast.
For BLIP, we observe a clear progression in attention behavior across layers.
In the early layers, the model attends broadly to the overall visual context, capturing global structures and trends.
As the layers deepen, attention becomes increasingly concentrated on specific peak regions that are most relevant to the target time-series prediction.
This transition from global context modeling to focused, task-relevant regions suggests that UniCast progressively refines visual information, filtering out irrelevant details and emphasizing salient cues for forecasting.
In contrast, CLIP exhibits a different attention pattern (shown in the appendix).
Early layers tend to focus on regions outside the plotted signal, while attention gradually shifts toward the plot area in later layers.
Despite these differences, both backbones demonstrate a consistent trend of aligning visual attention with regions relevant to time-series prediction in deeper layers, indicating that UniCast effectively guides pretrained vision models toward task-specific reasoning.

\paragraph{Text}
To understand how UniCast utilizes textual information, we visualize attention heatmaps over text tokens together with the corresponding forecasting outputs.
As shown in Figure~\ref{fig:textheatmap}, when the text description contains trend-related phrases such as “upward trend” or “downward trend”, the model assigns higher attention weights to these tokens.
This attention pattern indicates that UniCast identifies semantic cues describing temporal dynamics from the text modality.
Consistent with this observation, the forecasting outputs exhibit trend behaviors aligned with the textual descriptions, producing predictions that follow upward or downward trajectories accordingly.
These results suggest that the text modality contributes high-level temporal guidance, particularly for modeling global trends in time-series forecasting.
Overall, the visualization demonstrates that UniCast effectively links semantic trend information in text descriptions to forecast dynamics.

\paragraph{Time Series}
To understand how UniCast utilizes temporal information during forecasting, we visualize attention heatmaps over time-series tokens.
Figure~\ref{fig:timeseriesheatmap} reveal that the model does not exclusively focus on the last observed token, which is commonly emphasized in autoregressive forecasting models.
Instead, UniCast selectively attends to multiple temporally relevant tokens, including intermediate positions that appear informative for prediction.
This behavior suggests that the model identifies salient temporal patterns and dependencies across the sequence, rather than relying solely on the most recent observation.
Such selective attention over time supports more effective temporal context modeling, particularly when important signals occur earlier in the sequence.

\section{Parameter Efficiency}
\label{app:parameterefficiency}

Table~\ref{tab:parameterefficiency} presents the parameter efficiency of different UniCast configurations compared to the time-series backbone Chronos.
This includes variations with and without Conditional Prompting (CP) and Modality Routing (MR), highlighting the impact of these components on model size.
\end{document}